\documentclass[10pt,journal,cspaper,compsoc]{IEEEtran}
\pdfoutput=1
\usepackage{amsfonts}
\ifCLASSOPTIONcompsoc
\usepackage[nocompress]{cite}
\else
\usepackage{cite}
\fi
\usepackage{amsmath}
\usepackage{algorithm}
\usepackage{algorithmic}
\usepackage{stfloats}
\usepackage{url}
\usepackage{mathrsfs}
\usepackage{dsfont}
\usepackage{multirow}
\usepackage{multicol}
\usepackage{color}
\usepackage{xcolor}
\usepackage{graphics}
\usepackage{graphicx}
\usepackage{subfig}
\usepackage{booktabs}

\usepackage[framemethod=tikz]{mdframed}
\usepackage{soul}
\soulregister{\cite}7 
\soulregister{\citep}7 
\soulregister{\citet}7 
\soulregister{\ref}7 
\soulregister{\textbf}7
\usepackage{cancel}

\begin{document}
	
	\title{Context Disentangling and Prototype Inheriting for Robust Visual Grounding}

	\author{Wei Tang, Liang Li, Xuejing Liu, Lu Jin, Jinhui Tang and Zechao Li
		\IEEEcompsocitemizethanks{
  \IEEEcompsocthanksitem W. Tang, L. Jin, J. Tang and Z. Li are with School of Computer Science and Engineering, Nanjing University of Science and Technology, No. 200 Xiaolingwei Road, Nanjing 210094, China. E-mail: \{weitang, zechao.li, lu.jin, jinhuitang\}@njust.edu.cn (Corresponding author: Zechao Li)
  \IEEEcompsocthanksitem L. Li is with Institute of Computing Technology, Chinese Academy of Sciences, Beijing 100190, China. E-mail: liang.li@ict.ac.cn
  \IEEEcompsocthanksitem X. Liu is with SenseTime Research, Beijing 100084, China. E-mail: liuxuejing@sensetime.com
  }
  
  }


\IEEEcompsoctitleabstractindextext{
\begin{abstract}
Visual grounding (VG) aims to locate a specific target in an image based on a given language query. 
The discriminative information from context is important for distinguishing the target from other objects, particularly for the targets that have the same category as others. However, most previous methods underestimate such information. Moreover, they are usually designed for the standard scene (without any novel object), which limits their generalization to the open-vocabulary scene. In this paper, we propose a novel framework with context disentangling and prototype inheriting for robust visual grounding to handle both scenes. 
Specifically, the context disentangling disentangles the referent and context features, which achieves better discrimination between them. The prototype inheriting inherits the prototypes discovered from the disentangled visual features by a prototype bank to fully utilize the seen data, especially for the open-vocabulary scene. The fused features, obtained by leveraging Hadamard product on disentangled linguistic and visual features of prototypes to avoid sharp adjusting the importance between the two types of features, are then attached with a special token and feed to a vision Transformer encoder for bounding box regression. Extensive experiments are conducted on both standard and open-vocabulary scenes. The performance comparisons indicate that our method outperforms the state-of-the-art methods in both scenarios. 
{The code is available at https://github.com/WayneTomas/TransCP.}
\end{abstract}
		
}

\maketitle
\IEEEdisplaynotcompsoctitleabstractindextext
\IEEEpeerreviewmaketitle

\section{Introduction}
\begin{figure}[t]
    \centering
    \includegraphics[width=1.0\linewidth]{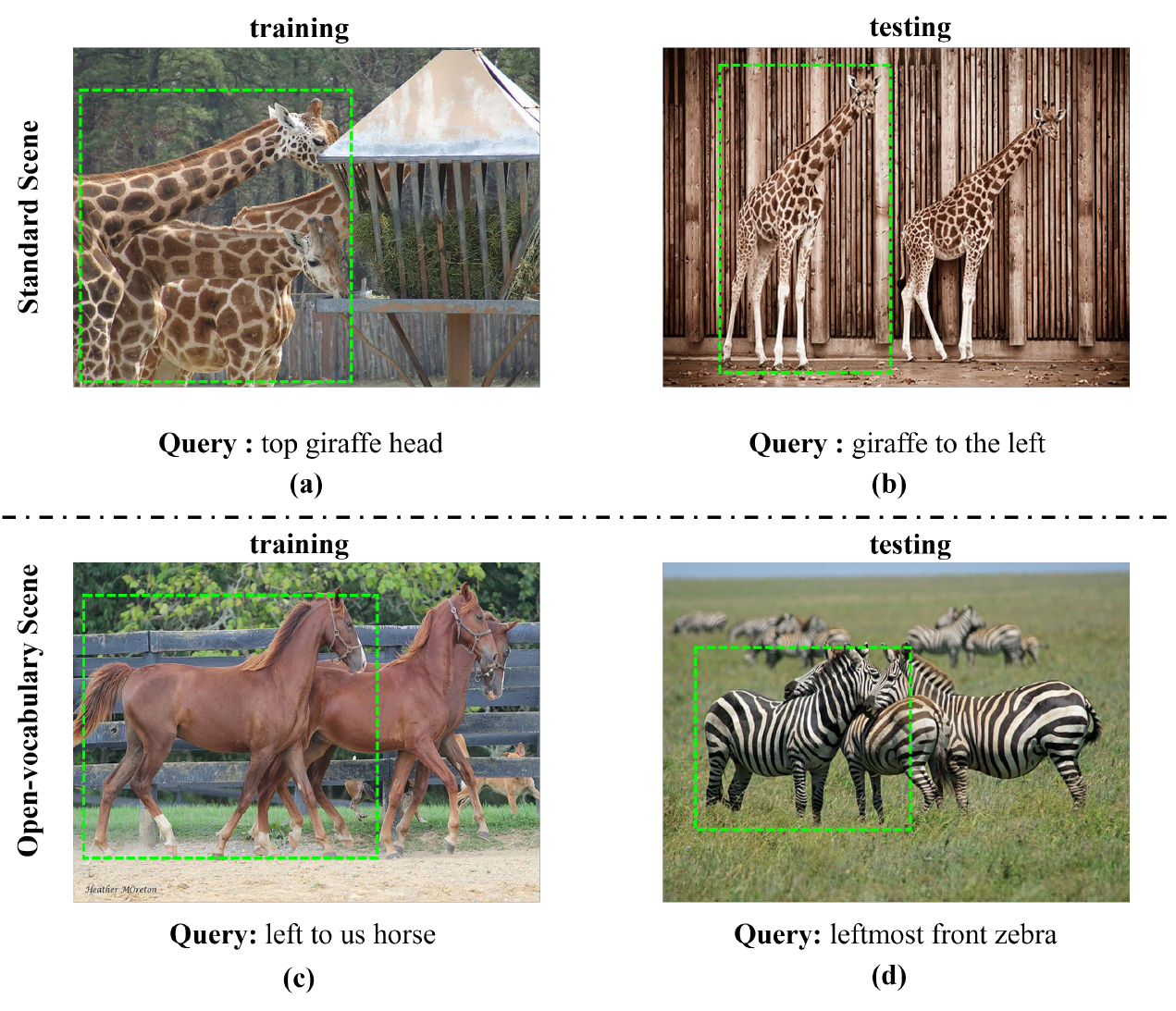}
    \caption{An illustration of the two different scenes in visual grounding. Examples (a) and (b) depict a standard scene where the concept "giraffe" appears both in the training data (a) and the testing data (b). However, in the open-vocabulary scene, the concept "zebra" in the testing data (d) falls outside the scope of the "horse" concept in the training data (c). It is of a cluster-level prototype, for example, "animal", exists in both training and testing data.}
    \vspace{5pt}
    \label{fig:introduce}
\end{figure}

\begin{figure*}[t]
    \centering
    \includegraphics[width=1\linewidth]{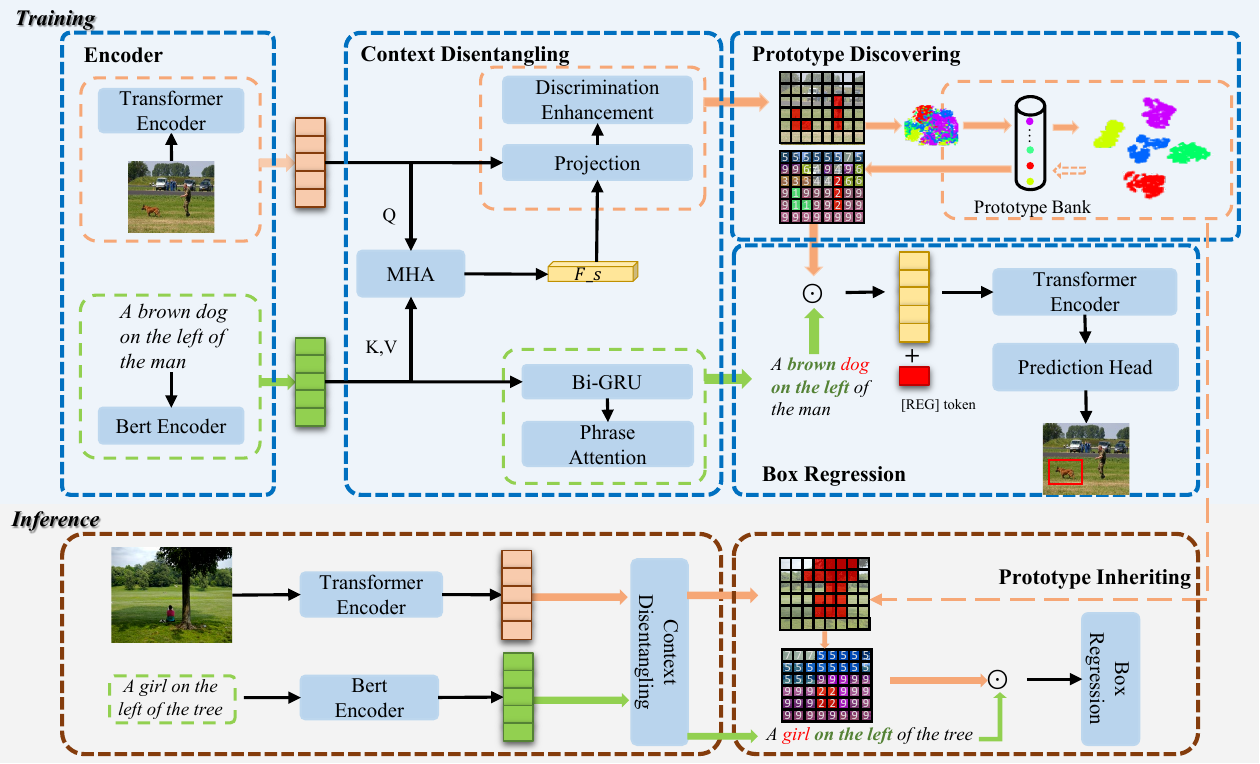}
    \caption{An illustration of robust visual grounding based on Transformers (TransCP) that disentangles context and inherits prototypes. Given an image and a query expression, the multi-modal features are extracted and disentangled. The prototypes in the prototype bank are discovered from the disentangled visual features during the training phase and then inherited during the inference phase. Finally, the boxes are regressed by feeding the fused visual and linguistic features attached with a special token to a vision Transformer encoder.}
    \label{fig:framework}
\end{figure*}

\IEEEPARstart{V}{isual} 
{Grounding (VG) \cite{liu2019adaptive, yang2022improving, chen2022understanding, du2022visual} aims to locate a specific object or region in an image referred by a natural language expression as shown in Fig.~\ref{fig:introduce}. This referred object or region is called referent, and it typically serves as the subject in the language expression.}
It is a fundamental task to realize the cognitive interaction between humans and machines \cite{liu2022entity}, such as visual question answering \cite{HuangHGQZ19}, social image understanding \cite{LiTIP17} and robot navigation \cite{yu2020language}. The goal of visual grounding is to enable computers to better understand human language by connecting it to visual data.

Previous visual grounding methods can be roughly divided into three categories, namely two-stage approaches, anchor-based one-stage approaches, and anchor-free one-stage approaches (i.e., Transformer-based). Pioneer two-stage methods \cite{CMN2017,VC2018,ParalAttn2018,yu2018mattnet,similar_net2019} learn the bounding box of referent by the cross-modal matching between a set of region proposals (e.g. Fast-RCNN) and the language features. In contrast, anchor-based one-stage methods \cite{chen2018real,sadhu2019zero,yang2019fast,liao2020real,yang2020improving,huang2021look} yield the bounding box according to the maximal confidence over dense anchors and the fused two-modality features. The recent anchor-free Transformer-based methods directly generate the bounding box by changing the grounding problem to a coordinates regression task, which is relieved from the pre-defined region proposals and anchors \cite{du2022visual,zhu2022seqtr,yang2022improving,deng2021transvg}. 
{However, previous methods still have room for improvement in learning discriminative information from the context that can better separate the referent and other context objects. Following \cite{yang2022improving}, we divide the context into two types, namely the visual context and the language context. We consider everything except the referent in the image as visual context while everything except the referent in the expression, including categories, attributes, relations, and relational objects (context objects) as language context. 
Effectively utilizing the discriminative information within the context can lead to better performance of visual grounding, especially when the referent has the same category as the context objects \cite{yang2022improving, YangLY19, liu2019adaptive, liu2022entity}.}

Furthermore, most previous methods are designed for standard scene (without any novel object), thus they are hard to generalize into open-vocabulary scene (base and novel objects are mixed together), which is illustrated in Fig.~\ref{fig:introduce}.
Recent works try to introduce external knowledge, such as auxiliary datasets \cite{liu2020transferrable} or multi-modal knowledge graphs \cite{ShiSJZ22} to learn the knowledge about the novel objects. However, the more novel objects are added, the more external knowledge is needed. Some other works like MDETR \cite{MDETR_KamathSLSMC21}, UniTab \cite{Unitab_YangGW000LW22}, and OFA \cite{WangYMLBLMZZY22} explore large-scale Vision-Language pre-training models for better generalize visual grounding in open-vocabulary scene. Nevertheless, these models require significant amounts of computational resources and can be challenging to train and fine-tune. Due to the large number of parameters and complex architectures, fine-tuning entire model on task-specific datasets might even damage the well-learned features.
As such, alternative approaches to visual grounding are needed that can better leverage the underlying structure and patterns in the data. Based on findings from cognitive psychology and neuroscience research \cite{rosch1978principles, murphy2002big} , grounding an object, whether novel or not, based on coarse-grained cluster-level information, i.e. prototypes~\cite{xu2022attribute, li2020transferrable}, is an inherent prior for human beings. However, previous methods usually ground the referent from fine-grained visual and linguistic features, thus are limited in the open-vocabulary scene.

Inspired by these observations, we propose a robust visual grounding framework TransCP to handle both standard and open-vocabulary scenes by disentangling context and inheriting prototypes. We first enhance the discrimination for referent both in the image and text. Then the prototypes are discovered and inherited from the disentangled visual features through a prototype bank, which makes full use of the seen data to ground the referent in both scenarios. 

As shown in Fig.~\ref{fig:framework}, TransCP consists of four modules including a two-branch encoder, a context disentangling module, a prototype discovering and inheriting module, and a box regression module. 
{Specifically, the context disentangling module is leveraged to disentangle the visual and linguistic features from the two-branch encoder for better distinguishing between the referent and the context.
In the visual branch, the salient objects (usually nouns appearing in the referring expressions, containing the referent object) can be distinguished with significant differences from other objects or regions through the cross-attention mechanism and discrimination enhancement.}
{In the text branch, the context information is emphasized via a phrase attention module by adaptively assigning higher soft weights to the context than the referent.}
{Second, the prototype discovering and inheriting module condenses scattered fine-grained visual semantic information to cluster-level prototypes, which captures the essential characteristics of the objects and stores in a prototype bank.} 
During the inference phase, the prototypes are inherited to help grounding the referents, especially for the novel objects in the open-vocabulary scene.
{Such visual and linguistic features are then fused by a parameter-free Hadamard that contributes to robustness. Unlike the parameter-depend multi-modal fusion strategies, the Hadamard fusion is more suitable for our method. With this strategy, the disentangled linguistic information acts as an attention mask to further enhance and select semantic-related visual features, which contributes to robustness. Furthermore, it avoids making sharp adjustments to the well-disentangled visual and linguistic features.}
Finally, in terms of bounding box prediction, we model it as a box regression problem by feeding the fused features attached with a special token to a vision Transformer encoder. 

In summary, our contributions are shown as follows:
\begin{itemize}
\vspace{-2mm}
    \item We propose a novel framework for robust visual grounding that deals with both standard and open-vocabulary scenes, which can handle both scenes more general and flexible.
    \item We design a context disentangling module to disentangle the referent and context, which better enhances the discrimination of the two types of features.

    \item We introduce a prototype inheriting module to condense the scattered semantic to cluster-level prototypes. They are inherited during the test phase for both scenes, especially for open-vocabulary.
\end{itemize}
	
\section{Relate Work}
	
In this section, we briefly review several related works about the visual grounding task in standard and open-vocabulary scene.

\subsection{Standard Scene}
Visual grounding aims to locate an object with respect to a given language referring expression. In crude terms, visual grounding approaches can be roughly divided into three types: two-stage methods, one-stage anchor-based methods, and one-stage anchor-free methods. The last type refers to Transformer-based methods currently.

\noindent\textbf{Two-stage methods.} Two-stage methods rely on a series of region proposals that are extracted from a pre-trained object detector. These early methods function like image-text cross-modal retrieval methods, matching the most appropriate region proposal to the referring expression \cite{nagaraja2016modeling, mao2016generation, luo2017comprehension, li2013partial, hu2016natural}. For instance, Spatial Context Recurrent ConvNet (SCRC) estimates the similarity scores among the multi-scale features (e.g. language features, local and global visual features) \cite{hu2016natural}. The authors in \cite{liu2019adaptive} and \cite{liu2022entity} model the weakly-supervised visual grounding as a cross-modal matching problem with the help of adaptive reconstruction and margin-based loss inspired by MAttnNet \cite{yu2018mattnet}. With the expansion of multi-modal big data, a range of Vision-Language Pre-training models have been suggested to deal with the visual grounding task. ViLBERT is proposed in \cite{lu2019vilbert}. It consists of a visual and language branch with Transformer and co-attention Transformer layers. ViLBERT first pre-trains the model on the set of paired visual proposals and annotations with the mask multi-modal modeling and multi-modal matching. Then the trained model is fine-tuned to adapt downstream tasks including visual grounding, which achieves much better performance than traditional methods. Different from ViLBERT of two branches, VL-BERT \cite{su2019vl} handles the vision and language in a unified framework. Besides, it only pre-trained on the mask multi-modal modeling task. 
{Some methods focus on exploring the complicated relations~\cite{LiTM19} in the visual and linguistic scenes with Graph Convolutional Networks (GCNs) \cite{YangLY20ref_reasoning, YangLY19, ZhengWTZCWW20, HuRDS19, DengWWHLT22ATT_A}. For example, Modular Graph Attention Network (MGA-Net) achieves promising performance in complex visual relational reasoning \cite{ZhengWTZCWW20}. A novel accumulated attention (A-ATT) and noise training mechanism is presented to reason among all the attention modules jointly to reduce internal redundancies  \cite{DengWWHLT22ATT_A}. It explicitly captures the relation among different kinds of information, which improves the performance and robustness of visual grounding.}
Although these two-stage methods promote the performance of visual grounding, they are constrained by the limitations of pre-trained object detectors.

\noindent\textbf{One-stage anchor-based methods.} To break the limitations of two-stage methods, there have been many one-stage methods. The exploration of the one-stage methods starts with anchor-based methods. SSG is the first attempt of one-stage methods in the field of visual grounding \cite{chen2018real}, where the authors propose a multi-modal interactor based on attention to produce the fused CNN and LSTM features for the referring expression grounder. FAOA based on YOLOv3 is presented in \cite{yang2019fast}. Darknet \cite{redmon2017yolo9000} is used as the backbone for extracting visual features while BERT is used for obtaining the language features. Spatial features are fed into the model to boost the grounding performance. These multi-modal features are then fused and input into the YOLOv3-based network for bounding box prediction. Their next work \cite{yang2020improving} suggests that recursive sub-query construction, along with multi-round fusion and reasoning, can address the problem that FAOA can not handle: the long and complex referring expressions. RCCF is proposed in \cite{liao2020real} that deems the grounding problem as a correlation filtering. The grounding result is the object center according to the highest value of the correlation map between visual and the given language features. A multi-task collaborative network is proposed to learn Referring Expression Grounding (REG) and Referring Expression Segmentation (RES) at the same time \cite{luo2020multi}. The authors find that RES will be conducive to multi-modal matching in the REG task. To better deal with the complex expression, an iterative shrinking strategy base on reinforcement learning is proposed \cite{sun2021iterative}. LBYL-Net \cite{huang2021look} fuses multi-scale features by landmark feature convolution module. The grounding module is based on YOLOv3 as same as FAOA. However, the tedious details of the fusion model and their anchor-based grounding model in these methods restrict their grounding ability.

\noindent\textbf{One-stage anchor-free methods.} To liberate one-stage grounding methods from the pre-defined dense anchors, researchers propose many anchor-free one-stage methods. They usually rely on Transformers that directly regress the coordinates of the bounding box based on the given expression. TransVG is the first attempt that introduces Transformers to visual grounding \cite{deng2021transvg}. The visual and linguistic features are extracted by DETR and BERT respectively, which are then concatenated with a learnable special token and fed to a Vision-Language Transformer for bounding boxes regression.
VGTR is proposed for one-stage visual grounding \cite{du2022visual}. Especially, a text-guided self-attention module is proposed to replace the cross-modal attention module in the Transformer encoder. The method achieves decent performance. The authors in \cite{zhu2022seqtr} propose SeqTR that solve REG and RES task in a unified framework. They treat REG and RES as a sequences prediction problem which can be seen as an extension of Pix2Seq. 
{RefTR is proposed in \cite{LiS21reftr}. It is an approach for Multi-task visual grounding, which jointly trains REG and RES in the same framework using two different task-specific heads. Moreover, RefTR supports the pre-training strategy for further improvements.}
Work in \cite{yang2022improving} proposes a context modeling framework for visual grounding with a multi-stage Transformer decoder. Specifically, the visual-linguistic verification module and language-guided context encoder are the keys to model the context, which improves the discrimination between the referring object and the context. 
{Pseudo-Q is proposed to construct the visual grounding model in an unsupervised manner, which leverages the generated pseudo query as the referring expression for supervised training \cite{JiangLHSH22Pseudo_Q}.}
Although these methods have achieved promising performance for visual grounding, they underestimate the discriminative information implicated in the context. Moreover, they are mostly designed for the standard scene and can hardly generalize to the open-vocabulary scene.

\subsection{Open-vocabulary Scene}
The open-vocabulary scene, in which some objects (novel objects) in testing data are not present in training data, is much more prevalent in practice. The seen and unseen data are mixed in this situation, and it is also called the general zero-shot in some literature \cite{0007CGPYC21}.

The authors suggest that the concepts can be transferred from an auxiliary classification data and the category-independent context \cite{liu2020transferrable}. The concepts can be inherited for helping the open-vocabulary grounding. ZSGNet for Zero-Shot grounding is proposed in \cite{sadhu2019zero}, which requires enough discrimination among the categories of queries. It depends on an anchor generator that generates dense anchor proposals. However, its performance on the standard scene is not satisfactory. The authors suggest to use external knowledge to help the zero-shot visual grounding \cite{ShiSJZ22}. The multi-modal knowledge graphs are explored to learn the knowledge about the novel objects.
In practice, we also find that some Transformer-based visual grounding methods achieve decent performance in the open-vocabulary scene, such as \cite{deng2021transvg},
\cite{du2022visual},
\cite{zhu2022seqtr} and
\cite{yang2022improving}.
  
Recently, large-scale Vision-Language pre-training (VLP) shows promising performance in the open-vocabulary scene. Some works like MDETR \cite{MDETR_KamathSLSMC21}, UniTab \cite{Unitab_YangGW000LW22}, and OFA \cite{WangYMLBLMZZY22} explore VLP models for visual grounding in open-vocabulary scene. Nevertheless, these models are high computational costs and can be challenging to train and fine-tune.
Some other works about open-vocabulary scene for different tasks can be find in image captioning \cite{hu2021vivo, li2020oscar}, object detection and recognition \cite{zareian2021open, TANG2022108792}, 
video classification \cite{qian2022multimodal}, meta learning \cite{TangLPT20}, etc.

Generally, most of these methods focus on introducing external knowledge or expanding the model scale, while underestimating the internal context information. However, some important discriminative information that distinguishes the referent from other regions is implicit in the context features. Furthermore, the cluster-level prototypes that can help to deal with both standard and open-vocabulary scenes are also ignored, which could better improve the robustness of the visual grounding model.
	
\section{The Proposed Approach}
\subsection{Overview}
Given an Image $\mathcal{I}$ and a linguistic expression query $\mathcal{Q}$, the goal of visual grounding is to predict a bounding box for the referent expressed in the language. In TransCP, we try to regress a sequence $\mathcal{B}=\{b_{i}\}$ of four-dimensional discrete bounding box parameters directly. Most previous methods focus on the standard scene. When they meet the open-vocabulary scene, which can also be regarded as a general zero-shot problem \cite{0007CGPYC21}, they tend to give poor performance. We devote our effort to handling the standard and open-vocabulary visual grounding in the same framework. As depicted in Fig.~\ref{fig:framework}, our proposed TransCP contains four main components, namely a two-branch encoder, a context disentangling module, a prototype discovering and inheriting module, and a box regression module.

\subsection{{Two branch Encoder}}\label{sec:enc}
Unlike the two-stage or anchor-based one-stage approaches that rely on pre-defined region features, models with limited pre-defined regions may have limited representation ability. Thus, for our visual branch, an cooperative trainable encoder, which constitutes the fundamental component of our end-to-end pipeline, is considered. While the CNN backbone focuses on local features, the later combined Transformer encoder pays more attention to global features with the help of Multi-head attention and position encoding modules. Specifically, the whole image $\mathcal{I}$ is first fed into a ResNet backbone to extract 2D visual feature maps $F_v^{C \times H \times W}$. Then, a stacked Transformer encoder is leveraged to tell the model pays more attention to global features. Finally, the visual features with local and global properties are denoted as ${F'}_v$. 
 \begin{equation}
    \begin{aligned}
        {F'}_v = E_v(\mathcal{I}, \theta_v) \in \mathbb{R}^{C \times HW},
     \end{aligned}
\end{equation}
where $E_v(\cdot, \theta_v)$ is the visual encoder stated above. $\theta_v$ denotes the encoder parameters.
Similarly, for the language branch, followed by other works, we use the 12-layer BERT as our referring expression encoder. Through the encoder, language sequence $\mathcal{Q}$ are encoded as a sequence of linguistic features $F_l$.
 \begin{equation}
    \begin{aligned}
        F_l = E_l(\mathcal{Q}, \theta_l) \in \mathbb{R}^{C \times L},
     \end{aligned}
\end{equation}
where $L$ is the length of the textual sequence, and $\theta_l$ denotes the parameters in the language encoder.

 \subsection{Context Disentangling}\label{sec:RC}
{One of the crucial problems of visual grounding is how to enhance the discrimination between the referent and the context. To achieve this purpose, we propose a context disentangling module. It is made up of two sub-modules for the visual and language branches respectively. The visual context disentangling enhances such discrimination indirectly. Specifically, we enhance the salient objects' features (containing referent object) and suppresses the context features with the help of cross-modal attention and discrimination coefficients. We refer to this process as visual context disentangling.}
Meanwhile, the language context disentangling module reaches a similar purpose that focuses the context information (e.g. relations, attributes) by leveraging the phrase attention after the adapted language features.
{We refer to this process as language context disentangling.}
 
 For the visual context disentangling module, we propose to leverage cross-modal attention to preliminary screening all salient objects that appear in the referring expression. The semantic features ${F}_s$ are first aggregated by the guidance of language features. Specifically, we use the visual feature ${F'}_v$ as the query and the language features $F_l$ as the key and value to get the semantic features ${F}_s$:
  \begin{equation}
    \begin{aligned}
        F_s = softmax(\frac{(W_Q^T{{F'}_v})(W_K^T{F_l})^T}{\sqrt{dim_K}})\cdot (W_V^T{F_l}),
     \end{aligned}
\end{equation}
where $W_Q$, $W_K$ and $W_V$ are the projection weights for query, key and value in the attention mechanism respectively. ${dim_K}$ denotes the channel dimension of the key.
{${F}_s$ can be regarded as the shared semantics between the visual and language features (to some extent, they represent the salient objects in the language features). Once we obtain the semantic features ${F}_s$, our goal is to identify the salient objects in the visual features. Specifically, the semantic features ${F}_s$ and visual features ${F'}_v$ are projected into the same metric space to measure their discrimination coefficients for further enhancing their discrimination.}
 \begin{equation}
    \begin{aligned}
        \mathcal{E} = \alpha \cdot \exp(-\frac{(1-S({{F'}_v},{F_s}))^2}{2\delta^2}), \label{enhance coefficients}
     \end{aligned}
\end{equation}
where $S(\cdot)$ is a similarity metric that we use a simple projection and a dot product in our modal. $\alpha$ and $\delta$ are two learnable parameters, respectively.
 The discrimination coefficients are weighted on the visual features ${F'}_v$ pixel-wisely to emphasize the salient objects (see the patches with 60\% transparency in Fig.~\ref{fig:framework} before the prototype bank or the red and green output patches with 60\% transparency from the prototype bank). 
 \begin{equation}
    \begin{aligned}
        \widetilde{F_v} = {{F'}_v} \cdot \mathcal{E},
     \end{aligned}
\end{equation}

{To be clearer, with the Gaussian-like form of Equation~(\ref{enhance coefficients}), if the visual features ${F'}_v$ are related to the semantic features ${F}_s$, their coefficients are higher than those dissimilar with a sharp gap. Therefore, the discrimination between the referent (containing in the salient objects) and the context in visual features is enhanced. This makes the model more sensitive to salient objects with higher coefficients while suppressing the context with lower coefficients.}

For the language context disentangling module, we tackle the language features in a similar way. 
{However, in this branch, the context information contained in the language features is emphasized with higher soft weights, while referent information is preserved by assigning lower attention weights. Therefore, the discrimination between the referent and the context in language features is enhanced.}
In practical terms, after the pre-trained BERT, a Bi-GRU network is leveraged to further adapt the language features $F_l$ extracted by the pre-trained BERT to the seen data. 
 \begin{equation}
    \begin{aligned}
        {F_l}' = [\overset{\rightarrow}{{F_l}'}, \overset{\leftarrow}{{F_l}'}] = {E_l}'(F_l, {\theta_l}') \in \mathbb{R}^{C \times L},
     \end{aligned}
\end{equation}
where ${F_l}'$ denotes the adapted language features that concatenate bidirectional outputs from the Bi-GRU network. ${E_l}'$ and ${{\theta_l}'}$ are the Bi-GRU encoder and its parameters, respectively. Then, the adapted language features ${F_l}'$ are used as the inputs to the phrase attention that assigns soft attention weights to each word as following.
 \begin{equation}
    \begin{aligned}
        \phi = softmax(FC({F_l}')),
     \end{aligned}
\end{equation}
where $FC(\cdot)$ is a fully connected layer. The phrase attention is randomly initialized and supervised by the paired training data. When we get the attention weights, they will be weighted to the language features.
 \begin{equation}
    \begin{aligned}
        \widetilde{F_l} = {F_l}' \cdot \phi,
     \end{aligned}
\end{equation}

In summary, the phrase attention re-considers the different importance of each word in the language features. The context is highlighted by the higher attention weights while the referent is preserved by assigning lower weights, leading to an enhancement of the discrimination between the referent and the context.

\subsection{Prototype Discovering and Inheriting}
The features that come from the visual branch are scattered and contain amounts of low-level features \cite{huang2021soho}. Sometimes these low-level features would be harmful to visual grounding since language expressions often manifest as high-level and abstract information. In contrary, the cluster-level information, i.e., prototypes, are helpful for both scenes since prototypes reflect the essential properties of the objects and can act as the coarse-grained guidance. Thus, we propose a prototype discovering and inheriting module that utilizes a prototype bank. With the prototype bank, we can discover cluster-level prototypes from the scattered visual features during the training phase. These prototypes are then stored in the bank and inherited during the inference phase to assist with standard and open-vocabulary visual grounding. The colored patches in Fig.~\ref{fig:framework}, labeled with the prototypes' identifiers, represent the quantized visual features obtained by leveraging the prototypes. Specifically, the prototype bank we used is defined as follows:
 \begin{equation}
    \begin{aligned}
        \mathcal{P}=\{p_{i}\}, i=1...k,
     \end{aligned}
\end{equation}
where $k$ is the number of prototypes and each prototype is a embedding vector with $c$ dimension. Next, the scattered fine-grained visual features are assigned to these cluster-level prototypes:
 \begin{equation}
    \begin{aligned}
        {\widetilde{F_v}}'=p_{j},
     \end{aligned}
\end{equation}
where $j$ is the assigned index. It can be calculated by the following formula:
 \begin{equation} \label{dicmapping}
    \begin{aligned}
        j=\min dist(\widetilde{F_v}, \mathcal{P}),
     \end{aligned}
\end{equation}
where $dist(\cdot)$ is a distance metric and we use the Euclidean distance since it is widely used to measure the distance between two visual features. Actually, the strategy we used is a nearest neighbor searching. However, the prototypes are discrete. The way that selects the minimal distance from the prototype bank is not differentiable and can not be optimized with the the gradient back propagation. To solve this problem, following \cite{van2017neural}, we adopt their optimizing method by using the stopping gradient strategy. Concretely, it can be optimized by reformulated Equation~(\ref{dicmapping}) as
 \begin{equation}
    \begin{aligned}
        {\widetilde{F_v}}'=sg[p_j-\widetilde{F_v}] + \widetilde{F_v},
     \end{aligned}
\end{equation}
where $sg[\cdot]$ is the operation that stop the gradient on the parameters. The gradient for previous parameters are transmitted by the auxiliary term $\widetilde{F_v}$, which makes the prototype bank trainable.

In analogy to most deep-learning-based clustering methods, our prototype bank also needs to update the prototypes within a mini-batch. The conventional clustering methods, e.g. $k$-means, update the cluster centroids by calculating the average over the data assigned to the centroids. Following \cite{huang2021soho}, We adopt the momentum average optimization as a replacement for the direct calculation of the average of visual features $\widetilde{F_v}$ assigned to the prototypes. The features' averages are updated as the new prototypes if the number of visual features that belong to the prototypes is not equal to zero.

Generally, the prototype bank performs mini-batch deep-learning-based clustering on the visual features. Through it, the visual features with the same or similar semantics are aggregated together. The scattered visual features are then transformed by leveraging cluster-level prototypes. Such conversion will benefit the cross-modal matching since the prototypes reflect the essential properties of the object and are closer to the linguistic features from the perspective of semantic scale. Besides, the prototypes stored in the prototype bank are inherited to guide the standard and open-vocabulary scenes during the inference phase, which works as same as a look-up table.

\subsection{Bounding Box Regression}
When we obtain the quantized visual features ${\widetilde{F_v}}'$ and the disentangled language features $\widetilde{F_l}$, the next matter is how to fuse these features to get multi-modal features for the bounding box regression. Traditional approaches employ complicated fusion methods which sometimes ignore the balance of visual and linguistic features. For example, a multi-scale feature fusion module is leveraged in \cite{yang2019fast}. \cite{yang2020improving} adopts a multi-stage cross-modal Transformer decoder that mulls over the visual and linguistic information iteratively. A Vision-Language Transformer is applied to integrate the cross-modal features with the multi-head attention \cite{deng2021transvg}. Different from these methods, we propose to use a simple yet effective Hadamard product as the fusion strategy. The disentangled linguistic information acts as an attention mask to further enhance and select semantic-related visual features, which contributes to robustness. Besides, the parameter-free Hadamard fusion strategy maintains the equal importance between the disentangled linguistic features and prototypes of disentangled visual features, which avoids drastic adjustments of the well-learned features.
 \begin{equation}
    \begin{aligned}
        F_m = tanh({\widetilde{F_v}}') \odot tanh(\widetilde{F_l}),
     \end{aligned}
\end{equation}
where $F_m$ is the fused multi-modal features.

Inspired by TransVG \cite{deng2021transvg}, our architecture also follows the design of a learnable special embedding (i.e. the [REG] token) for bounding box regression. Different from Vision-Language Transformer encoder utilized in TransVG that performs intra- and inter-modality relation reasoning through the concatenation of visual and language tokens, we leverage only a vision Transformer encoder as an alternative solution. This decision is motivated by the fact that the multi-modal features have already been derived through the use of the Hadamard fusion. In addition, because the regression will be affected by the variance shifting problem with dropout layers \cite{A2020Effect}, we cut down these layers by setting their dropout probabilities equal to zero in the regression Transformer encoder. 
 \begin{equation}
    \begin{aligned}
        {[REG]}_{output} = E_{r}([REG]_{initial}, F_m], \theta_{r}){[0]} \in \mathbb{R}^{C \times 1},
     \end{aligned}
\end{equation}
where $E_{r}$ is the vision Transformer encoder and $\theta_{r}$ is the corresponding weights. Besides, the [REG] token is initialized randomly and concatenated to the head of the $F_m$.
The output state of the [REG] token is then used as the input of the bounding box regression head. The regression head consists of an MLP layer with three layers and two corresponding ReLU active functions. The output of it is a sequence $\mathcal{B}$ organized as $(x, y, w, h)$ that means the coordinates of the top left vertex, the width, and the length for the regressed bounding box.
 \begin{equation}
    \begin{aligned}
        \mathcal{B} = MLP([REG]_{output}),
     \end{aligned}
\end{equation}
\subsection{Training and Inference}
Different from conventional visual grounding methods that rely on the region proposals or pre-defined anchors, our framework straightly regresses a sequence that contains necessary parameters for determining a bounding box in the image with the given referring expression. Following other Transformer-based methods, the loss function applied to the training phase of our model is the widely used L1 and IoU loss.
 \begin{equation}
    \begin{aligned}
        \mathcal{L} = \lambda_{l1} \mathcal{L}_{l1}(B, \hat{B}) + \lambda_{GIoU} \mathcal{L}_{GIoU}(B, \hat{B}),
     \end{aligned}
\end{equation}
where $b$ and $\hat{b}$ denote the ground truth bounding box and the regressed bounding box, respectively. $\lambda_{l1}$ and $\lambda_{GIoU}$ are two trade-off factors that balance the two losses which are set to $5$ and $2$ empirically. Moreover, in the inference stage, the prototype bank is frozen that works like a look-up table. Through the prototype bank, the cluster-level prototypes are inherited to help grounding the referent in both scenarios, especially for open-vocabulary scene.
	
\section{Experiment}
\subsection{Datasets}
\noindent\textbf{ReferIt.} ReferIt \cite{kazemzadeh2014referitgame}, also known as ReferCLEF, contains 20000 labeled images.
This dataset includes some uncertain queries, such as "any", "whole" and some wrong-labeled data. Following previous works \cite{deng2021transvg, yang2022improving}, three subsets are split for training, validation, and testing, which consists of 54,127, 5,842, and 60,103 queries respectively.

\noindent\textbf{Flickr30k Entities.} Each image in Flickr30k Entities \cite{PlummerWCCHL17} is attached with five refer expressions. 31783 images and 427k short referred phrases rather than long sentences are contained in this dataset. Moreover, it does not focus on images with multiple objects of the same category.

\noindent\textbf{RefCOCO/RefCOCO+/RefCOCOg.}
The images of these datasets all come from MSCOCO \cite{lin2014microsoft} dataset. RefCOCO contains 50,000 objects from 19,994 images with 142,209 referring expressions. It is split officially into training, validation, testA, and testB subjects with 120,624, 10,834, 5,657 and 5,095 expressions, respectively. For the testA, the expressions mostly referred the people while in the testB they mainly related to various objects except for people. RefCOCO+ contains 141,564 referring expressions attached to 49,856 objects from 19,992 images. Unlike RefCOCO, the absolute locate expression, such as "right", "top" and "besides", are forbidden in RefCOCO+. As a result that RefCOCO+ pays more attention to the appearance attributes of referents. Training, validation, testA, and testB are split officially with 120,191, 10,758, 5,726, and 4,889 expressions respectively. RefCOCOg has 25,799 images, 49,822 objects with 95,010 corresponding referring expressions. There are two split versions of this dataset, i.e., RefCOCOg-google \cite{mao2016generation} and RefCOCOg-umd \cite{nagaraja2016modeling}. To simplify the open-vocabulary evaluation in section~\ref{comparision_open}, we only report the results on RefCOCOg-google (val-g) following \cite{feng2021encoder}, \cite{yang2020improving}, \cite{huang2021look} \cite{zhu2022seqtr} and \cite{liu2019adaptive}.

\noindent{\textbf{Ref-Reasoning.}}
{Ref-Reasoning \cite{YangLY20ref_reasoning} is a large-scale real-world dataset built on GQA \cite{HudsonM19gqa} for structured referring expression reasoning, which consists of 791,956 referring expressions in 83,989 images. The referring expressions are automatically generated based on scene graphs from the Visual Genome dataset \cite{KrishnaZGJHKCKL17visualgenome} by a set of expression templates. It contains 1,664 object classes, 308 relation classes, and 610 attribute classes. Consequently, the generated expressions are extremely complex, with each expression describing a maximum of 5 objects and their relationships. The dataset is divided into training, validation, and testing sets, comprising 721,164, 36,183, and 34,609 image-expression pairs, respectively. It is important to note that, to validate the generalization of the proposed method when confronted with extremely complex expressions, we conducted additional evaluations exclusively on Ref-Reasoning, as presented in Section~\ref{complex_dataset}.}

\subsection{Implementation Details} \label{imple_details}
\noindent\textbf{Data pre-processing.} We follow \cite{deng2021transvg} and \cite{yang2022improving} that resize each input image to $640 \times 640$, while the original aspect ratio is maintained. The longer edge will be resized and the shorter will be padded. For referring expression, we set the max query length to 20 including the special token [CLS] and [SEP]. Similarly, the longer and shorter ones will be cutted and padded respectively.

\noindent\textbf{Metric.} {Following recent works \cite{deng2021transvg, JiangLHSH22Pseudo_Q, yang2022improving, LiS21reftr}, the top-1 accuracy (\%) is adopted as our evaluation metric. The regressed bounding box is considered correct if its IoU with the ground truth box is larger than 0.5.}

\noindent \textbf{Training details.} We train the model with AdamW optimizer. 
{The initial learning rate is set to 1e-4 for our model, excluding the two-branch feature encoder, for which we set its learning rate to 1e-5.} 
The weights of the CNN backbone and the Transformer encoder are initialized by DETR model \cite{carion2020end}. For the learnable parameters $\alpha$ and $\sigma$ in Eq. \ref{enhance coefficients}, we empirically set their initial values to $1$ and $0.5$, respectively. Xavier initialization is utilized as the strategy to initialize the other parameters in our model. We train RefCOCO/RefCOCO+/RefCOCOg for $100$ epochs. For other datasets, 90 epochs are empirically set. Furthermore, we decay the learning rate by $10$ after $60$ epochs. For avoiding the cold start problem in the prototype discovering and inheriting module, i.e. the wrong optimizing direction caused by the randomly initialize prototypes, we freeze the CNN backbone for the first $10$ training epochs. Because we treat the visual grounding problem as a coordinates regression problem, the variance shifting problem existed in dropout layers \cite{A2020Effect} should be noted. As a solution, we set the dropout probabilities equal to zero in the bounding box regression module. Following \cite{yang2022improving}, we set the tradeoff coefficients in the loss function to $\lambda_{l1} = 5$ and $\lambda_{GIoU} = 2$ empirically.

\noindent \textbf{Inference.} {We train our model on the standard scene and select the model according to the best validation accuracy for testing.} For example, we train and validate on ReferIt dataset. For the testing phase, we separate it into two different parts. First, we test the model on standard datasets. Second, we test it for validating the robustness of open-vocabulary grounding. For simplicity of open-vocabulary testing, we train the model 1) on ReferIt and test on RefCOCO/RefCOCO+/RefCOCOg and Flickr30k entities; 2) on RefCOCO and test on ReferIt and Flickr30k; 3) on Flickr30k and test on RefCOCO/RefCOCO+/RefCOCOg and ReferIt; since this cross dataset testing exactly fits definition of the open-vocabulary scene that including both base and novel objects.

\subsection{Comparison Methods}
In order to evaluate the performance of our TransCP, we compare with different kinds of visual grounding methods. 

\textbf{Two-stage methods}: 
{CMN \cite{CMN2017}, VC \cite{VC2018}, ParalAttn  \cite{ParalAttn2018}, MAttnNet \cite{yu2018mattnet}, Similarity Net \cite{similar_net2019}, CITE \cite{CITE2018}, DDPN \cite{DDPN2018}, LGRANs \cite{LGRANs2019}, DGA \cite{DGA2019}, RvG-Tree \cite{RvG_Tree2019}, A-ATT \cite{DengWWHLT22ATT_A}, NMTree \cite{NMTree2019} and Ref-NMS \cite{Ref-NMS2021}}.

\textbf{One-stage anchor-based methods}: SSG \cite{chen2018real}, ZSGNet \cite{sadhu2019zero}, FAOA \cite{yang2019fast}, RCCF \cite{liao2020real}, ReSC-Large \cite{yang2020improving} and LBYL-Net \cite{huang2021look}.

\textbf{Transformer-based anchor-free methods}: 
{VGTR \cite{du2022visual}, Pseudo-Q \cite{JiangLHSH22Pseudo_Q}, RefTR \cite{LiS21reftr}, SeqTR \cite{zhu2022seqtr}, VLTVG \cite{yang2022improving} and TransVG \cite{deng2021transvg}}.

{Further, LBYL-Net \cite{huang2021look}, RefTR \cite{LiS21reftr}, VGTR \cite{du2022visual}, SeqTR \cite{zhu2022seqtr}, VLTVG \cite{yang2022improving}, and TransVG \cite{deng2021transvg} used as the compared methods of open-vocabulary scene are retrained in our environments with the same random seed, batch size, and PyTorch version, which mark with the symbol $\dagger$. The unsupervised method Pseudo-Q \cite{JiangLHSH22Pseudo_Q} is marked with the symbol $*$. Especially, VGTR and SeqTR are adapted with the appropriate size of language embedding layer for open-vocabulary testing.}

\subsection{Comparisons in the Standard Scene}
\begin{table}[t]
  \centering
  \caption{Comparison with the state-of-the-art methods on the test sets of ReferIt and Flickr30k Entities. In addtion, RN, DN and DLA are the abbreviation of ResNet, DarkNet and Deep Layer Aggregation, respectively. For example, ResNet-101 is abbreviated to RN101.
}
\setlength\tabcolsep{8pt}
  {
    \begin{tabular}{c|cccc}
    \toprule
          Method & Backbone & ReferIt test & Flickr30k test \\
    \midrule
    Two-stage: &     &    &       \\
    CMN &VGG16 & 28.33 & -  \\
    VC & VGG16 & 31.13 & - \\
    MAttnNet & RN101 & 29.04 & - \\
    Similarity Net & RN101 & 34.54 & 60.89 \\
    CITE & RN101 & 35.07 & 61.33 \\
    DDPN & RN101 & 63.00 & 73.30 \\
    \midrule
    
    One-stage: &&& \\
    SSG & DN53 & 54.24 & -\\
    ZSGNet & RN50 & 58.63 & 63.39 \\
    FAOA & DN53 & 60.67 & 68.71 \\
    RCCF & DLA34 & 63.79 & -\\
    ReSC-Large & DN53 & 64.60 & 69.28 \\
    LBYL-Net$^\dagger$ & DN53 & 68.14 & - \\
    \midrule
    Transformer-based: &&& \\
    Pseudo-Q $^*$ & RN50 & 43.32 & 60.41 \\
    RefTR $^\dagger$ & RN50 & 69.04 & 74.75 \\ 
    VGTR $^\dagger$ &RN50 & 63.63 & 75.44\\
    SeqTR $^\dagger$ & DN53 & 66.97 & 76.26 \\
    VLTVG$^\dagger$ & RN50 & 71.41 & 78.62 \\
    TransVG$^\dagger$ & RN50 & 69.86 & 78.07 \\
    \midrule
    TransCP &RN50 & \textbf{72.05}& \textbf{80.04} \\
    \bottomrule
    \end{tabular}%
    }
  \label{tab:Referit_Flickr30k_base}%
\end{table}%

\begin{table*}[t]
  \centering
  \caption{Comparison with the state-of-the-art methods on the test sets of RefCOCO, RefCOCO+ and RefCOCOg.}
  \setlength\tabcolsep{15pt}
  {
    \begin{tabular}{c|cccccccc}
    \toprule
    & &\multicolumn{3}{@{}c@{}}{RefCOCO} & \multicolumn{3}{@{}c@{}}{RefCOCO+} &\multicolumn{1}{@{}c@{}}{RefCOCOg}\\
    Method & Backbone & val &  testA  & testB & val & testA & testB & val \\
    \midrule
    Two-stage: &      & &   &   &   & &      \\
    CMN &VGG16 & - & 71.03 & 65.77 & -  & 54.32  & 47.76 & 57.47 \\
    VC & VGG16 & - & 73.33 & 67.44 & - & 58.40 & 53.18 & 62.30 \\
    ParalAttn & VGG16 & - & 75.31 & 65.52 & - & 61.34 & 50.86 & 58.03\\
    MAttnNet & RN101 & 76.65 & 81.14 & 69.99 & 65.33 & 71.62 & 56.02 &- \\
    LGRANs & VGG16 & - & 76.60 & 66.40 & - & 64.00 & 53.40 & 61.78\\
    DGA & VGG16 & - & 78.42 & 65.53 & - & 69.07 & 51.99 & - \\
    RvG-Tree & RN101 & 75.06 & 78.61 & 69.85 & 63.51 & 67.45 & 56.66 & -\\
    A-ATT & VGG16 & - & 80.87 & 71.55 & - & 65.13 & 55.01 & 63.84 \\
    NMTree & RN101 & 76.41 & 81.21 & 70.09 & 66.46 & 72.02 & 57.52 & 64.62\\
    Ref-NMS & RN101 & 80.70 & 84.00 & 76.04 & 68.25 & 73.68 & 59.42 & -\\
    \midrule
    One-stage: &      & &   &   &   & &      \\
    SSG & DN53 & - & 76.51 & 67.50 & - & 62.14 & 49.27 & 47.47 \\
    FAOA & DN53 & 72.54 & 74.35 & 68.50 & 56.81 & 60.23 & 49.60 & 56.12 \\
    RCCF & DLA34 & - & 81.06 & 71.85 & - & 70.35 & 56.32 & -\\
    ReSC-Large & DLA34 & 77.63 & 80.45 & 72.30 & 63.59 & 68.36 & 56.81 & 63.12\\
    LBYL-Net$^\dagger$ & DN53 & 79.74 & 82.71 & 73.84 & 68.96 & 73.52 & 59.44 & 63.78\\
    Transformer-based: &      & &   &   &   & & \\
    Pseudo-Q $^{*}$ & RN50 & 56.02 & 58.25 & 54.13 & 38.88 & 45.06 & 32.13 & 49.82 \\
    RefTR $^\dagger$ & RN50 & 80.92 & 83.40 & 75.78 & 69.22 & 74.61 & 60.95 & 61.76 \\
    VGTR $^\dagger$ &RN50 & 80.48 & 82.52 & 75.79 & 63.46 & 70.21 & 53.57 & 63.76 \\
    SeqTR $^\dagger$ & DN53 & 78.22 & 81.47 & 73.80 & 66.01 & 70.23 & 55.68 & 68.26\\
    VLTVG$^\dagger$ & RN50 & 83.21 & 86.78 & 78.45 & 72.36 & 77.21 & \textbf{64.80} & 71.40\\
    TransVG$^\dagger$ & RN50 & 80.48 & 82.57 & 75.79 & 66.22 & 71.25  & 57.39 & 67.41 \\
    \midrule
    TransCP &RN50 & \textbf{84.25} & \textbf{87.38} & \textbf{79.78} & \textbf{73.07} & \textbf{78.05} & 63.35 & \textbf{72.60} \\
    \bottomrule
    \end{tabular}%
    }
  \label{tab:COCO_base}%
\end{table*}%

In Table~\ref{tab:Referit_Flickr30k_base}, we report the top-1 accuracy (\%) of our TransCP and other state-of-the-art methods on test sets of ReferIt and Flickr30k Entities. 
To simplify the experiments in Section~\ref{comparision_open}, we only report the performances of the Transformer-based models with ResNet-50 (except SeqTR with DarkNet-53) backbone. The compared models are separated into three types, namely two-stage, one-stage (abbreviated for one-stage anchor-based methods), and Transformer-based (i.e., one-stage anchor-free) models.

We can observe that our model achieves the best performance on both datasets. Specifically, TransCP obtains $72.05\%$ top-1 accuracy on the ReferIt test set. From the perspective of two-stage models, our model significantly outperforms these models even though they may have a stronger visual backbone, such as MAttnNet, CITE, and DDPN. This phenomenon indicates that the two-stage methods built on pre-defined region proposals may not be optimized since their representation ability is restricted by the pre-trained object detectors. From the one-stage group, we can observe that our performance is absolutely $4.58\%$ higher than the strongest one-stage model LBYL-Net on the ReferIt dataset. 
{For the Transformer-based group, our method obtains consistent improvements over these methods. Specifically, our method surpasses Psedudo-Q, RefTR, VGTR, VLTVG, and SeqTR on the Flickr30k entities test set by using the same scale backbone with absolute accuracy of 19.63\%, 5.29\%, 4.60\%, 1.42\%, and 3.78\% respectively.
Compared with our baseline method TransVG, our TransCP achieves absolute improvements of up to 2.19\% and 1.97\%, which are remarkable margins on ReferIt and Flickr30k Entities.} 

To further validate our proposed framework, we conduct experiments on RefCOCO, RefCOCO+, and RefCOCOg respectively. 
{The results of our method compared with other state-of-the-art methods are shown in Table~\ref{tab:COCO_base}. From the results, it can be observed that our method consistently surpasses two-stage methods (A-ATT, NMTree, Ref-NMS, etc.) by a remarkable margin.}
Specifically, TransCP is higher than the strongest two-stage method Ref-NMS which leverages the stronger ResNet-101 as its backbone. The improvement is absolute $3.55\%$, $3.38\%$ and $3.74\%$ over all three splits set of RefCOCO (i.e. val, testA and testB). The reason for their sub-optimal results is the same as they rely on the pre-trained object detector which not involves in the training step of visual grounding. When the language expression becomes long and complicated, i.e., in RefCOCOg, our method shows a promising performance $72.06\%$, which surpasses all previous state-of-the-art methods. For one-stage methods, we can observe that the best LBYL-Net is lower than the strongest two-stage methods Ref-NMS on RefCOCO, but slightly higher than Ref-NMS on RefCOCO+. Surprisingly, our TransCP outperforms the best LBYL-Net on val, testA and testB of RefCOCO+ absolute $4.11\%$, $4.53\%$ and $3.91\%$ respectively. 
Compared with the Transformer-based methods, we can observe that they all surpass the previous one-stage methods (except the unsupervised method Pseudo-Q). It might be because Transformers provide more promising cross-modal interaction than the traditional YOLO-based one-stage methods. From the perspective of Transformer based methods, our method consistently achieves the best performance on RefCOCO and RefCOCOg datasets. When the RefCOCO+ forbids the absolute location expression, our method achieves $73.07\%$ and $78.05\%$ on val and testA respectively. The results surpass the most competitive method VLTVG. However, our TransCP is slightly lower than it on RefCOCO+ testB. This result suggests that there is still room for improvement in our method with regard to fully utilizing the discriminative information of the referent. Compared with our baseline TransVG, our approach achieves full improvement on RefCOCO, RefCOCO+, and RefCOCOg for all splits since we take full advantage of disentangled features. Besides, TransCP learns more representative features from the prototype discovering.

To sum up, the experimental results demonstrate the superiority of our method in the standard scene.

\subsection{Comparisons in the Open-vocabulary Scene} \label{comparision_open}
To more comprehensively understand the proposed TransCP, we conduct the experiments under the open-vocabulary settings. Specifically, we train the models in the standard scene and test them in the three different open-vocabulary scenes detailed in Section~\ref{imple_details}.
\begin{table*}[t]
  \centering
  \caption{Comparison of open-vocabulary settings with the state-of-the-art methods by the models trained on ReferIt dataset and test on RefCOCO/RefCOCO+/RefCOCOg and Flicker30k Entities.}
  \setlength\tabcolsep{15pt}
  {
    \begin{tabular}{c|ccccccccc}
    \toprule
    & \multicolumn{2}{@{}c@{}}{Flickr30k}&\multicolumn{3}{@{}c@{}}{RefCOCO} & \multicolumn{3}{@{}c@{}}{RefCOCO+} &\multicolumn{1}{@{}c@{}}{RefCOCOg}\\
    Method & val & test & val &  testA  & testB & val & testA & testB & val \\
    \midrule
    LBYL-Net$^\dagger$ & 26.00 & 26.19 & 55.61 & 61.75 & 46.26 & 38.03 & 43.14 & 29.29 & 40.02 \\
    RefTR$^\dagger$ & 29.88 & 30.47 &  58.16 & 61.15 & 52.83 & 36.49 & 40.15 & 32.67 & 44.96 \\
    VGTR$^\dagger$ & 14.31 & 15.01 & 10.03 & 7.25 & 12.91 & 11.1 & 7.75 & 13.60 & 8.44\\
    SeqTR$^\dagger$ & 17.64 & 18.38 & 9.96 & 9.61 & 10.13 & 11.05 & 9.78 & 11.82 & 5.76 \\
    VLTVG$^\dagger$ & 51.40 & 53.13 & \textbf{64.54} & \textbf{65.83} & \textbf{61.98} & \textbf{41.04} & \textbf{43.70} & \textbf{38.33} & 47.11\\
    TransVG$^\dagger$ & 52.29 & 54.38 & 61.67 & 63.23 & 59.39 & 37.52 & 39.13 & 35.10 & 45.79 \\
    \midrule
    TransCP & \textbf{52.92} & \textbf{55.07} & 64.26 & 65.55 & 61.71 & 40.17 &  42.70 & 36.90 & \textbf{47.40} \\
    \bottomrule
    \end{tabular}%
    }
  \label{tab:Open-vocab-referit}%
\end{table*}%

\begin{table}[t]
  \centering
  \caption{Comparison of open-vocabulary settings with the state-of-the-art methods by the models trained on RefCOCO dataset and test on ReferIt and Flicker30k Entities.}
  \setlength\tabcolsep{9pt}
  {
    \begin{tabular}{c|cccc}
    \toprule
    & \multicolumn{2}{@{}c@{}}{Flickr30k}&\multicolumn{2}{@{}c@{}}{ReferIt} \\
    Method & val & test & val &  test\\
    \midrule
    LBYL-Net$^\dagger$ & 21.76 & 22.93 & 22.97 & 21.62 \\
    RefTR $^\dagger$ & 27.67 & 29.45 & 24.03 & 23.08 \\
    VGTR$^\dagger$ & 23.11 & 23.46 & 9.57 & 9.17\\
    SeqTR$^\dagger$ & 28.78 & 29.53 & 15.45 & 13.97 \\
    VLTVG$^\dagger$ & \textbf{38.76} & 40.54 & 25.28 & 24.31 \\
    TransVG$^\dagger$& 36.11 & 37.86 & 23.78 & 22.83 \\
    \midrule
    TransCP & 38.35 & \textbf{40.62} & \textbf{29.01} & \textbf{27.71}  \\
    \bottomrule
    \end{tabular}%
    }
  \label{tab:Open-vocab-refcoco}%
\end{table}%

\begin{table*}[t]
  \centering
  \caption{Comparison of open-vocabulary settings with the state-of-the-art methods by the model trained on Flickr30k Entities dataset and test on RefCOCO/RefCOCO+/RefCOCOg and ReferIt.}
  \setlength\tabcolsep{15pt}
  {
    \begin{tabular}{c|ccccccccc}
    \toprule
    & \multicolumn{2}{@{}c@{}}{ReferIt}&\multicolumn{3}{@{}c@{}}{RefCOCO} & \multicolumn{3}{@{}c@{}}{RefCOCO+} &\multicolumn{1}{@{}c@{}}{RefCOCOg}\\
    Method & val & test & val &  testA  & testB & val & testA & testB & val \\
    \midrule
    RefTR$^\dagger$ & 36.61 & 35.70 & 29.72 & 36.16 & 23.21 & 27.67 & 31.85 & 22.21 & 35.13 \\
    VGTR$^\dagger$ & 8.59 & 8.49 & 18.33 & 17.35 & 11.85 & 9.09 & 5.78 & 11.95 & 13.05\\
    SeqTR$^\dagger$ & 11.80 & 11.83 & 12.41 & 12.52 & 11.38 & 12.70 & 12.96 & 11.63 & 18.44\\
    VLTVG$^\dagger$ & 44.13 & 42.25 & \textbf{39.67} & \textbf{46.33} & \textbf{31.82} & 36.28 & 41.83 & 31.34 & 38.63\\
    TransVG$^\dagger$ & 43.07 & 41.08 & 36.82 & 44.67 & 30.70 & 35.17 & 41.97 & 31.01 & 44.85\\
    \midrule
    TransCP & \textbf{45.36} & \textbf{43.99} & 39.35 & 46.26 & 31.76 & \textbf{37.80} & \textbf{42.91} & \textbf{32.95} & \textbf{44.91}\\
    \bottomrule
    \end{tabular}%
    }

  \label{tab:Open-vocab-flickr}%
\end{table*}%

In Table~\ref{tab:Open-vocab-referit}, we report the open-vocabulary performance with the model trained on ReferIt and tested on RefCOCO/RefCOCO+/RefCOCOg and Flickr30k Entities datasets. We can observe that TransCP achieves the best performance on Flickr30k and RefCOCOg datasets. However, compared with VLTVG, our method achieves comparable performance on RefCOCO and is slightly lower on RefCOCO+. This may be because the expression in ReferIt is very complex as it contains mis-labeded regions, and ambiguous queries such as "any" and "no". The low-quality context information may cause negative effects on prototype discovering and inheriting. Besides, relationships like "right" and "besides" are prohibited in RefCOCO+. 
{However, our model depends on the disentangled language features. If the discriminative information of relations is unavailable, it can also cause a negative impact on our model in the open-vocabulary scene.
This conclusion can also be summarized through the visualization in Fig.~\ref{fig:visualization_phrase_attn}. 
On the contrary, VLTVG only focuses on obtaining the salient features, which is less affected by low-quality expressions and non-working relationships.}

In Table \ref{tab:Open-vocab-refcoco}, we report the open-vocabulary performance with the model trained on RefCOCO and tested on ReferIt and Flickr30k Entities datasets. Our method achieves the best performance on the test split. Opposite to the model trained on ReferIt and tested on RefCOCO, TransCP exceeds VLTVG with a remarkable margin when we trained it on RefCOCO and test on ReferIt. This phenomenon demonstrates our hypothesis that a low quantity of training data would have a harmful effect on prototype learning, causing a negative impact in the open-vocabulary scenario.

 We can observe from Table~\ref{tab:Open-vocab-flickr} that our method has consistent improvements over the current state-of-the-art methods on ReferIt, RefCOCO+, and RefCOCOg. Besides, it achieves comparable performance on RefCOCO dataset compared with VLTVG. Compared with the baseline TransVG method, our method achieves more than 2\% absolute improvements. Moreover, as expressions in Flickr30k entities are short phrases and do not focus on images with multiple objects of the same category, simply enhancing the referent based on salient scores in VLTVG will not be effective when faced with long and complicated referring expressions in RefCOCOg.

{Furthermore, from Table~\ref{tab:Open-vocab-referit}, Table~\ref{tab:Open-vocab-refcoco} and Table~\ref{tab:Open-vocab-flickr}, we can find that VGTR and SeqTR  (LSTM features for language branch) have little generalization capability to handle the open-vocabulary scene, while LBYL-Net, RefTR, VLTVG, TransVG, and TransCP (BERT features for language branch) have the certain ability.} 
This phenomenon demonstrates that the good language features have a great significance in the open-vocabulary scene (of course in the standard scene) since the model needs to comprehensively understand which referent to identify in the image. Our ablation study in Section~\ref{ablation} also demonstrates the same conclusion.

\subsection{Comparisons on Ref-Reasoning} \label{complex_dataset}

\begin{table*}[t]
  \centering
  \caption{Comparison of open-vocabulary settings with the state-of-the-art methods by the models trained on Ref-Reasoning and test on Flickr30k Entities, ReferIt and RefCOCO.}
  \setlength\tabcolsep{17pt}
  {
    \begin{tabular}{c|ccccccc}
    \toprule
    &  \multicolumn{2}{@{}c@{}}{Flickr30k}&\multicolumn{2}{@{}c@{}}{ReferIt} & \multicolumn{3}{@{}c@{}}{RefCOCO} \\
    Method & val & test & val &  test & val & testA & testB \\
    \midrule
    VLTVG$^\dagger$ & 52.34 & 54.62 & 39.61 & 38.67 & 48.63 & \textbf{51.60} & 44.53 \\
    TransVG$^\dagger$ & 52.62 & 54.30 & \textbf{42.24} & 40.92 & 48.69 & 51.41 & 45.65 \\
    \midrule
    TransCP & \textbf{52.69} & \textbf{55.31} & 41.77 & \textbf{41.09} & \textbf{49.50} & 51.37 & \textbf{46.59} \\
    \bottomrule
    \end{tabular}%
    }

  \label{tab:ref_feasoning_Open-vocab}%
\end{table*}%

{To evaluate the generalization of TransCP when encountering extremely complex referring expressions in the open-vocabulary scene, we conduct experiments on the Ref-Reasoning dataset under the open-vocabulary settings. Specifically, we train TransCP on the Ref-Reasoning training set with the same settings as ReferIt. Then, we use the trained model to test it on the Flickr30k Entities, ReferIt, and RefCOCO datasets for open-vocabulary validation.}

{As shown in Table~\ref{tab:ref_feasoning_Open-vocab}, generally, TransCP outperforms the compared methods in most open-vocabulary scenarios. For example, the accuracy of TransCP exceeds VLTVG and TransVG by $0.69\%$ and $1.01\%$ on the Flickr30k test split, respectively.
Especially, for datasets containing a more diverse set of objects like RefCOCO testB split, our method achieves absolute $46.59\%$ that outperforms VLTVG with a significant margin ($+ 2.06\%$). The results indicate that the patterns of the disentangled features and the prototypes learned from training data are more suitable for the open-vocabulary scene than the patterns learned by other methods. 

Furthermore, we provide a comprehensive analysis by incorporating experiments trained on ReferIt and tested on RefCOCO (Table~\ref{tab:Open-vocab-referit}).
It is important to emphasize that ReferIt is also a complex dataset that is challenging to transfer to the open-vocabulary scene. The key distinction between ReferIt and Ref-Reasoning is that ReferIt contains numerous ambiguous expressions, whereas Ref-Reasoning comprises longer and more complex expressions. From Table~\ref{tab:ref_feasoning_Open-vocab} and Table~\ref{tab:Open-vocab-referit}, we find that the models trained on Ref-Reasoning and tested on Flickr30k Entities achieve similar performance to those trained on ReferIt and tested on Flickr30k Entities. Therefore, we consider the performance on Flickr30k Entities as an anchor so we can compare the result in Table~\ref{tab:ref_feasoning_Open-vocab} and Table~\ref{tab:Open-vocab-referit} simultaneously. Some observations can be drawn as follows:

1) When dealing with extremely complex expressions in the training set, it is more challenging to learn suitable information for the open-vocabulary scene from seen data. The open-vocabulary performance of the models (VLTVG, TransVG, and TransCP) trained on Ref-Reasoning and tested on RefCOCO decreased more than $15\%$ compared to those trained on ReferIt and tested on RefCOCO.

2) As the complexity of referring expressions increases, TransCP demonstrates more robustness than the compared methods. From Table~\ref{tab:Open-vocab-referit}, we can observe that VLTVG slightly outperforms our method on RefCOCO. However, from Table~\ref{tab:ref_feasoning_Open-vocab}, when the models are trained on Ref-Reasoning contained longer and more complex relations, our method achieves better open-vocabulary performance than VLTVG, such as the val split ($+ 0.83\%$) and the testB split ($+ 2.06\%$).
}

\subsection{Ablation Study} \label{ablation}
\begin{table*}[t]
  \centering
  \caption{The ablation study of different fusion strategies in the standard scene (RefCOCO) and the open-vocabulary scene (ReferIt). "V.L." and "H.D." are the abbreviations for Vision-Language Transformers and Hadamard fusion, respectively.}
\setlength\tabcolsep{17pt}
  {
    \begin{tabular}{cccccccc}
    \toprule
    \multicolumn{2}{@{}c@{}}{Fusion}&  \multicolumn{3}{@{}c@{}}{RefCOCO} & \multicolumn{2}{@{}c@{}}{ReferIt}\\
    V.L. & H.D. &  val & testA & testB & val & test\\
    \midrule
    $\surd$& & 83.02 & 86.41  & 79.04 & 25.54 & 24.24  \\
    & $\surd$ & \textbf{84.25} & \textbf{87.38} & \textbf{79.78} & \textbf{29.01} & \textbf{27.71} \\
    \bottomrule
    \end{tabular}%
    }
  \label{tab:hd_vl_ablation}%
  \end{table*}

\begin{table*}[t]
  \centering
  \caption{The ablation study of different modules in standard scene (RefCOCO) and open-vocabulary scene (ReferIt). "H.D.", "L.D.", "V.D." and "P.T." are the abbreviation for Hadamard fusion, linguistic disentangling, visual disentangling and prototype discovering and inheriting, respectively.
}
\setlength\tabcolsep{17pt}
  
  {
    \begin{tabular}{ccccccccc}
    \toprule
    &\multicolumn{1}{@{}c@{}}{Linguistic}&\multicolumn{2}{@{}c@{}}{Visual} & \multicolumn{3}{@{}c@{}}{RefCOCO} & \multicolumn{2}{@{}c@{}}{ReferIt}\\
    H.D. & L.D. & V.D. & P.T. &  val & testA & testB & val & test\\
    \midrule
    & & & & 80.48 & 82.57 & 75.79 & 23.78 & 22.83 \\

    &$\surd$& & & 82.46 & 85.35 & 77.96 & 24.46 & 23.55 \\
    &&$\surd$& & \textbf{84.47} & 86.44 & 79.04 & 25.50 & 23.99 \\
    &&&$\surd$& 81.48 & 84.25 & 75.41 & 23.90 & 22.69 \\
    
    \midrule
    $\surd$& & & & 82.90 & 86.02 & 77.90 & 25.47 & 24.19 \\
    $\surd$&$\surd$& & & 83.86 & \textbf{87.41} & 79.76 & 27.27 & 26.59 \\
    $\surd$&&$\surd$& & 84.36 & 86.65 & 79.31 & 27.71 & 26.51 \\
    $\surd$&&& $\surd$ & 83.55 & 85.89 & 78.59 & 26.16 & 24.45 \\
    $\surd$&$\surd$&$\surd$& & 83.90 & 87.04 & 79.41 & 28.53 & 27.11 \\
    $\surd$&$\surd$&$\surd$&$\surd$& 84.25 & 87.38 & \textbf{79.78} & \textbf{29.01} & \textbf{27.71} \\
    \bottomrule
    \end{tabular}%
    }
  \label{tab:ablation}%
  \end{table*}
{First of all, we invest the impact of different fusion strategies, including parameter-free Hadamard fusion and parameter-depend Vision-Language Fusion.}
Then to investigate the impact of different components of our model, we conduct the ablation study of the model trained on RefCOCO dataset. 
{Ablation studies are implemented on the Hadamard fusion, visual context disentangling, language context disentangling, and prototype discovering and inheriting modules. These components are abbreviated as "H.D.", "L.D.", "V.D." and "P.T." respectively.}
We also perform experiments on different sizes of the prototype bank and different layers of the vision Transformer encoder used in box regression stage in this section. 
{To further analyze the proposed TransCP, the comparison of the time cost during the training and inference phase is given.}

\subsubsection{Fusion Strategies}
\label{section_fusion}
{In Table \ref{tab:hd_vl_ablation}, we present the results of TransCP combined with different fusion strategies. Based on the results, we validate that H.D. used in TransCP contributes to robustness in both scenes. We can observe that TransCP (with H.D. fusion strategy) achieves about absolute $1\%$ higher than the one equipped with V.L. across all splits of RefCOCO in the standard scene. In the open-vocabulary scene, the advantage of TransCP with H.D. is even more pronounced. Specifically, TransCP with H.D. achieves absolute $29.01\%$ and $27.71\%$, respectively, which is more than absolute $3\%$ higher than the one with V.L. 
The reasons why H.D. contributes to robustness in both scenes may be as follows. Firstly, the disentangled linguistic information acts as an attention mask to further enhance and select semantic-related visual features, which contributes to robustness. The other reason is that the Hadamard fusion is a parameter-free strategy. Unlike V.L., which learns dynamic fusion importance between visual and linguistic features from training data, H.D. maintains equal importance between the two types of features. It avoids making sharp adjustments to the well-disentangled visual and linguistic features. }

\subsubsection{Components of TransCP}
In Table \ref{tab:ablation}, we present the results of TransCP with different modules on RefCOCO dataset. 
We can make several observations as follows. First, without any of the modules that we utilized in this paper, our model actually degenerates into the baseline model TransVG \cite{deng2021transvg}. 
{Next, we study the model combined with only L.D., only V.D., and only P.T., respectively. We can observe that with only L.D., V.D., and P.T., the performances are decent in the standard and open-vocabulary scenes. For example, with only L.D., the model achieves only about a $2\%$ improvement over the baseline. This phenomenon indicates that with only L.D. or V.D., the model still can not make full use of the discriminative information mined from disentangled features. These two modules are interdependent. V.D. enables the model to focus on a few of the most likely regions, while L.D. can leverage discriminative context information (relations, attributes, etc., see Section~\ref{sec: visual_context_disentangling} for more details) to further determine the true referent. When we focus on the model with only P.T., we observe that it achieves comparable performance to TransVG. This indicates that the effectiveness of P.T. depends on the previously disentangled visual features. These disentangled visual features ensure that the prototype discovering module can learn good prototypes, thereby enhancing the model's robustness in both scenes.}

{After introducing the Hadamard product fusion module, we can observe that the performance improved by more than two percent in both the standard scene (RefCOCO) and the open-vocabulary scene (ReferIt). These improvements may be because the adapted visual and linguistic features are already of high quality. Detailed comparisons of two different fusion strategies are discussed in Section~\ref{section_fusion}. Based on the results and analysis, it is better to fuse the two disentangled features with a parameter-free strategy before applying the box regression module to maintain their equal importance.}
Second, by leveraging language disentangling, the performance of TransCP has been further improved. For example, TransCP achieves an absolute improvemet of $1.86\%$ on RefCOCO testB dataset and $2.40\%$ on ReferIt test dataset. This phenomenon demonstrates that by disentangling language features and further enhancing the context, the discrimination between the referent and context is improved. 
{When we replace L.D. with V.D., the performance is comparable to the one with H.D. and L.D. in both scenes. For the case that with only V.D., we can find that H.D. contributes to open-vocabulary scene rather than standard scene. This further demonstrates that not only the prototype discovering module contributes to the generalization property, but also other designs including context disentangling module and Hadamard product fusion strategy do. Besides, the case of H.D. combined with P.T. also supports our findings.}
Third, to our surprise, adding the visual disentangling does not bring about many improvements in the standard scene. However, we are gratified to observe that visual disentangling can further boost the performance of the open-vocabulary scene by about absolute $1.26\%$ on the validation set and $0.52\%$ on the test set of the ReferIt dataset, respectively.  We infer from our experience that the language disentangling has given our model sufficient ability to learn abundant discriminative information from the standard scene dataset. Thus, while TransCP reaches a bottleneck in the standard scenario, adding visual disentangling may not be effective in overcoming it. In contrast, the performance of TransCP in the open-vocabulary scene does not exhibit such a bottleneck.
The disentangled visual features highlight the salient objects. Then, the disentangled linguistic features provide the highlighted context information, which helps the model better distinguish the novel referent from the context.
Lastly, the complete model that incorporates prototypes overcomes the aforementioned limitations. The cluster-level prototypes not only benefit standard scenes, resulting in an improvement of approximately $0.3\%$, but also achieve an additional absolute gain of $0.6\%$ on the ReferIt test dataset of the open-vocabulary scene. The prototypes mined from seen data can ease the semantic matching of visual and linguistic features in the Transformers for bounding box regression by bringing the scale of the two types of features closer. Furthermore, from the result of the complete model in open-vocabulary scene, it can be found that the inheritance of these prototypes can guide the downstream grounding task in both standard and open-vocabulary scenes.
\subsubsection{Size of Prototype Bank}
\begin{figure}[t]
    \centering
    \includegraphics[width=1\linewidth]{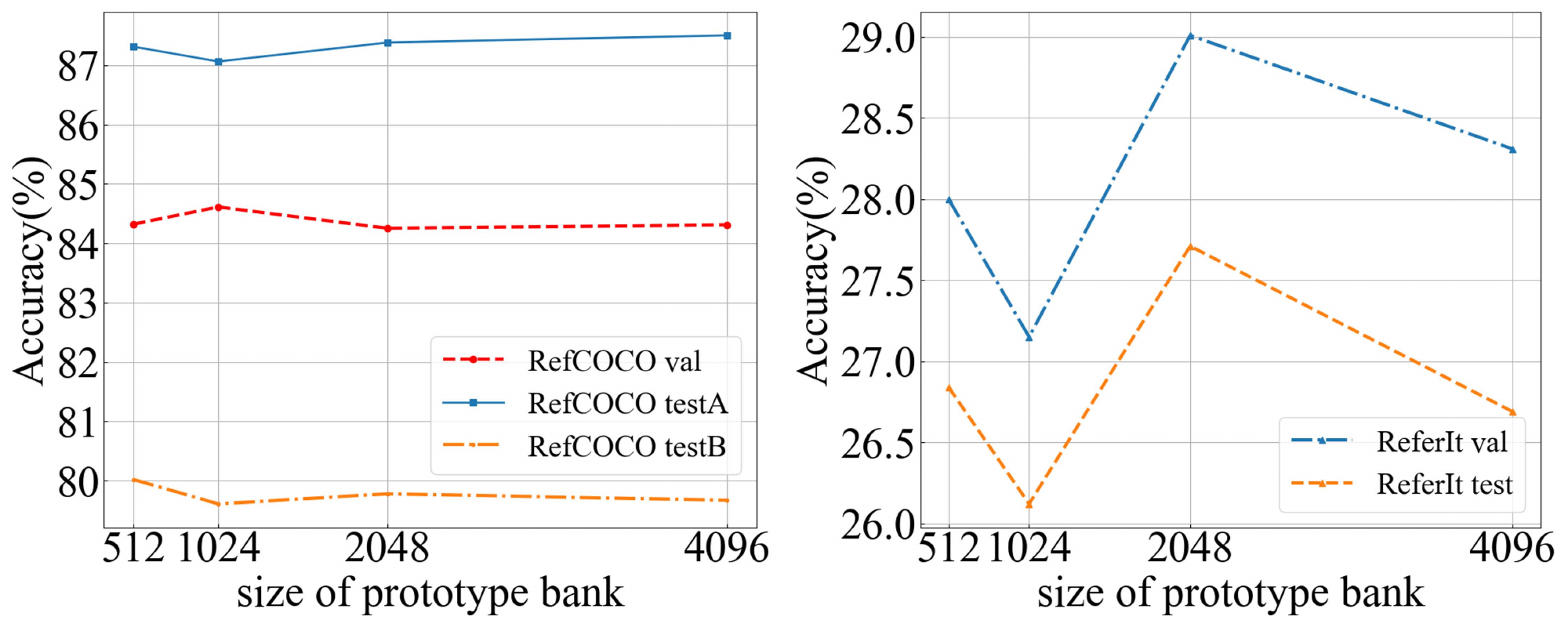}
    \caption{Ablation study for different size of prototype bank. We train the model on RefCOCO dataset and test in standard scene of RefCOCO and open-vocabulary scene of ReferIt.}
    \label{fig:Figure_ablation}
\end{figure}

Fig.~\ref{fig:Figure_ablation} shows the impact of different size of prototype bank. We can observe several things from the results. First, different sizes of prototype bank have little influence on the performance of our model for standard visual grounding. The size of prototype bank actually controls the coarseness of the prototypes that extracted from the seen data. Since the standard visual grounding scene holds limited categories, generally all prototypes can be learned from different coarseness with many epochs in the training phase. 
Second, when dealing with novel objects in the open-vocabulary scene, the performance of our model is considerably affected by the size of the prototype bank. 
A prototype bank that is either too large or too small does not yield satisfactory performance. The coarser the prototypes are, the greater the ambiguity and generality of the concept information will be, and vice versa. It indicates that when transferring these prototypes to deal with the open-vocabulary scene, we should well control the coarseness. To achieve the balance for both standard and open-vocabulary scenes, according to Fig.~\ref{fig:Figure_ablation}, we choose the size of $2048$ in our proposed model.

\subsubsection{Number of Transformer Encoder Layers}
\begin{figure}[t]
    \centering
    \includegraphics[width=1\linewidth]{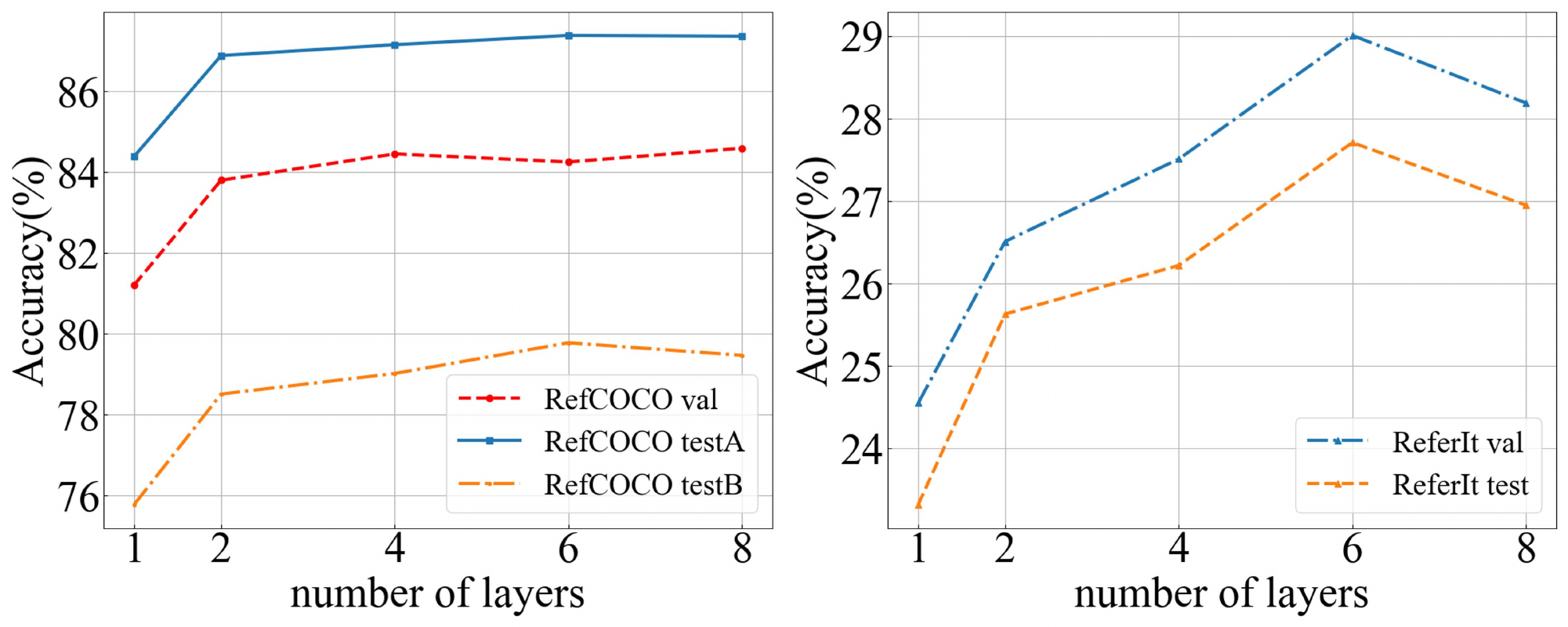}
    \caption{Ablation study for numbers of Transformer encoder layers used in bounding box regression. We train the model on RefCOCO dataset and test in the standard scene of RefCOCO and the open-vocabulary scene of ReferIt.}
    \label{fig:Figure_ablation_layers}
\end{figure}
Fig.~\ref{fig:Figure_ablation_layers} shows the impact of the different number of vision Transformer encoder layers in the bounding box regression stage. We can observe that the accuracy continuously increases on RefCOCO testA and testB of the standard scene when we add more layers. When the layers grow to a certain number of $6$, our model on testA and testB achieves the best performance. However, the accuracy slightly decreases when the number of Transformer layers becomes larger. In general, for our model in a standard scene, performance remains stable up to 6 layers. We can observe the other situation on the right of Fig.~\ref{fig:Figure_ablation_layers} that in the open-vocabulary scene, the number of layers has a great impact on the accuracy of ReferIt val and test set. Furthermore, we find that $6$ layers Transformer encoder is the best choice for our model. Considering both aspects of standard and open-vocabulary scenes, we employ a $6$-layer vision Transformer encoder in our model for bounding box regression.

\subsubsection{Time Cost}
\begin{table}[t]
  \centering
  \caption{Comparison of the time cost with the state-of-the-art methods on the testB split of RefCOCO dataset. 
}
  \setlength\tabcolsep{6pt}
  {
    \begin{tabular}{c|cccc}
    \toprule
    Method & Params. & Training (ms)  & Inference (ms) \\
    \midrule
    VLTVG$^\dagger$ & 151.26M & 358.23 & 79.70 \\
    TransVG$^\dagger$ & 149.52M & 281.98 & 74.32 \\
    \midrule
    TransCP w/o P.T. & 159.85M & 250.35 & 72.74 \\
    TransCP & 160.84M & 273.98 & 78.86 \\
    \bottomrule
    \end{tabular}%
    }
  \label{tab:time_consumption}%
\end{table}%
{To better analyze the proposed TransCP, we conduct an experiment to measure the time cost during both the training and inference phases. During the training phase, the time cost is recorded as the average time ($ms$) of $10$ randomly selected samples for forward and backward procedure. During the inference phase, only forward time cost is recorded. The comparison models VLTVG, TrasVG, TransCP without P.T. and TransCP are executed on a single GPU of GTX 3090 24GB with a batch size of $1$. We choose the RefCOCO testB split as the experimental dataset. The results are shown in Table~\ref{tab:time_consumption}.

It can be observed that the training cost of the proposed TransCP is the most economical overall compared to TransVG and VLTVG. The proposed TransCP takes approximately $273$ ms per sample during training. 
Without P.T., TransCP achieves a relative speed-up of approximately $9\%$ per sample during the training phase while reducing the number of trainable parameters by approximately $1M$. However, from Table~\ref{tab:ablation}, we can see that the complete model of TransCP surpasses TransCP w/o P.T. a relative accuracy of approximately $2\%$ in the open-vocabulary scene of the ReferIt test split.
In addition, the time cost during the inference phase and the count of trainable parameters are comparable across all the models under consideration. All these observations indicate that maintaining a prototype bank does not substantially increase the training costs. Furthermore, the performance can be significantly improved through this module in both standard and open-vocabulary scenes.}

\section{Qualitative Results}
\subsection{Visualization of Context Disentangling} 
\label{sec: visual_context_disentangling}

\begin{figure}[t]
    \centering
    \includegraphics[width=1\linewidth]{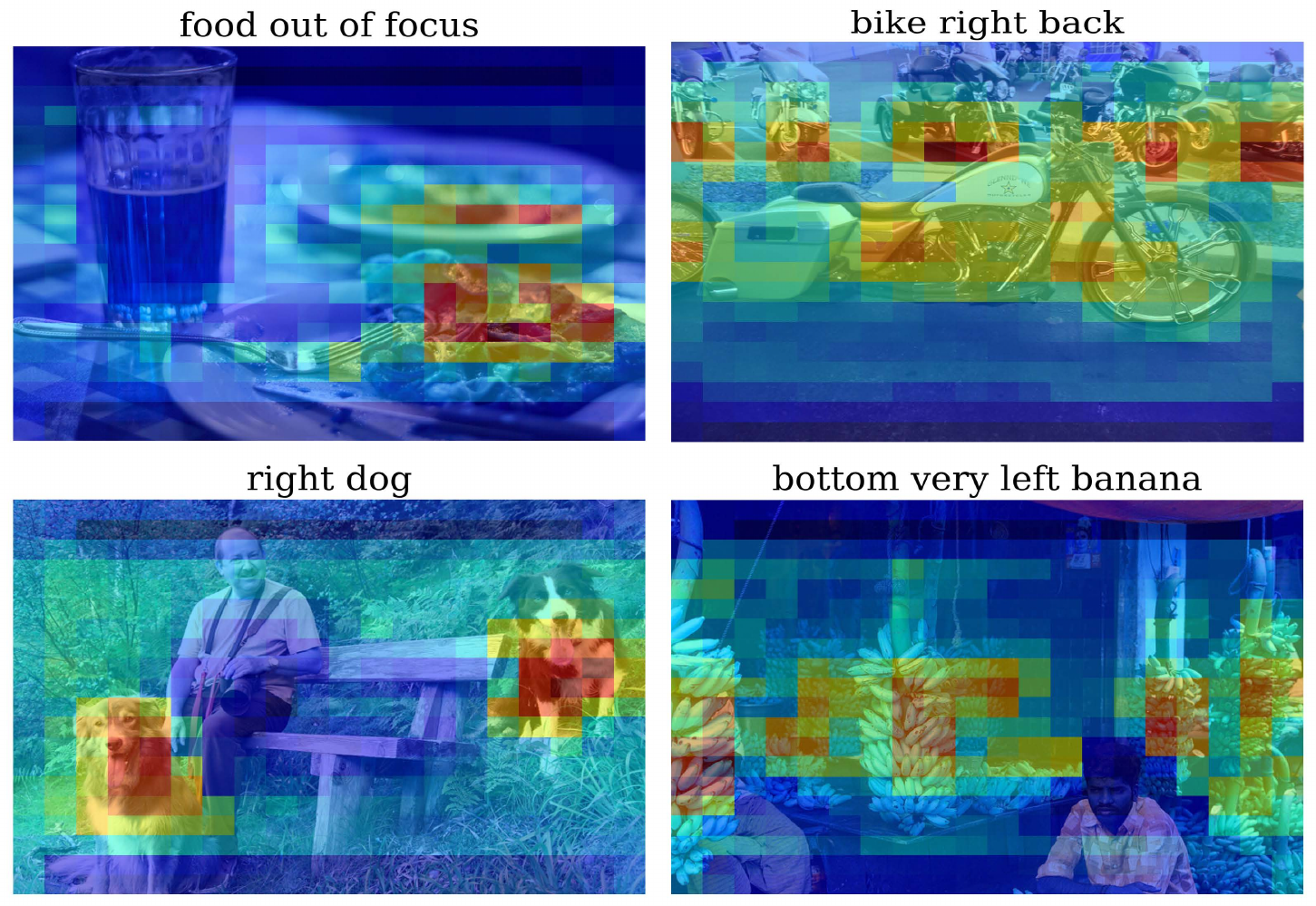}
    \caption{Qualitative results of discrimination coefficients in the visual branch of context disentangling module.}
    \label{fig:visualization_dis_score}
\end{figure}

\begin{figure}[t]
    \centering
    \includegraphics[width=1\linewidth]{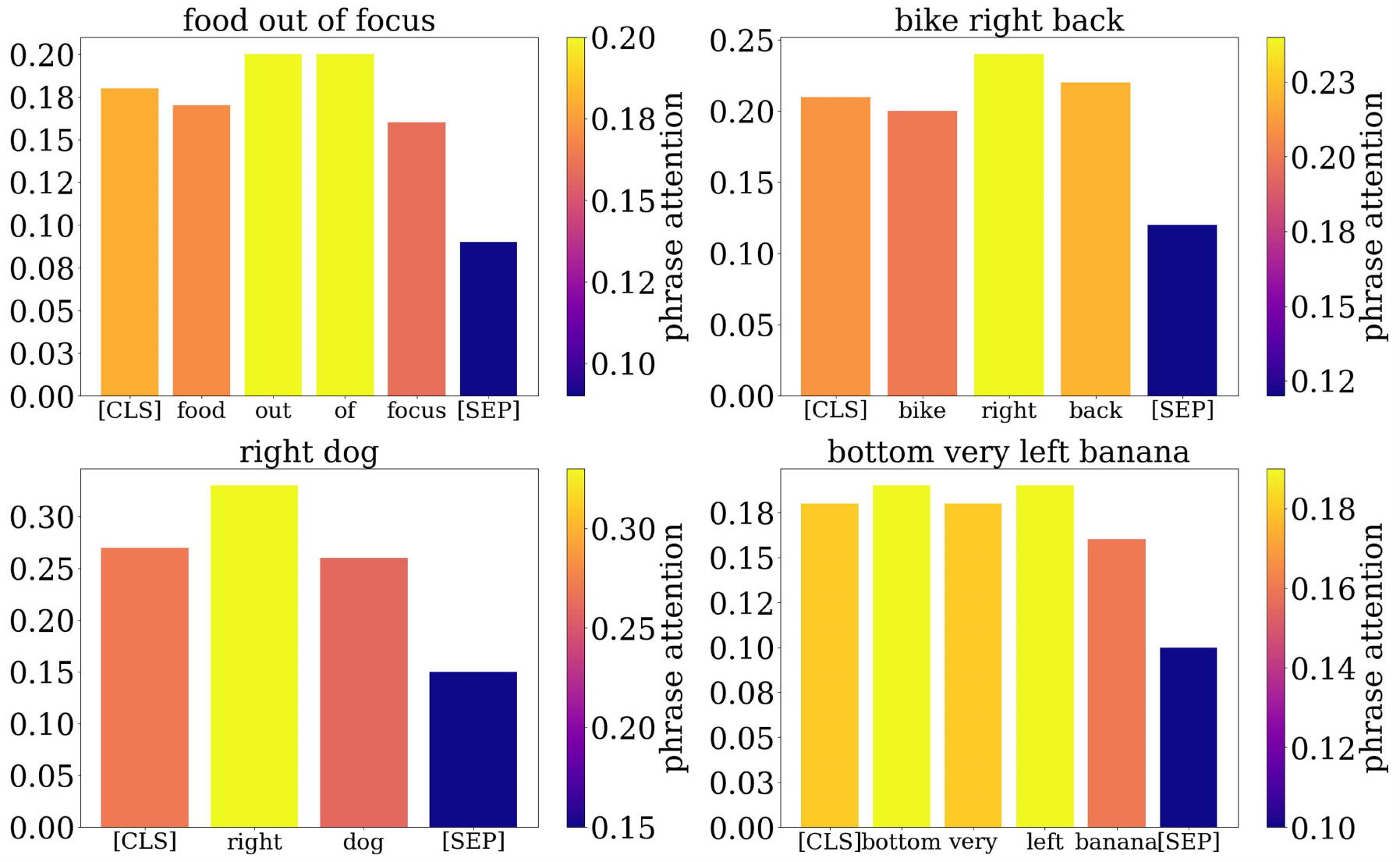}
    \caption{Qualitative results of phrase attention weights in the language branch of context  disentangling module.}
    \label{fig:visualization_phrase_attn}
\end{figure}

\definecolor{Red}{RGB}{255,0,0}
\definecolor{Green}{RGB}{17,204,34}
\definecolor{Brown}{RGB}{190.0, 129.0,65.0}
\begin{figure*}[t]
    \centering
    \includegraphics[width=1\linewidth]{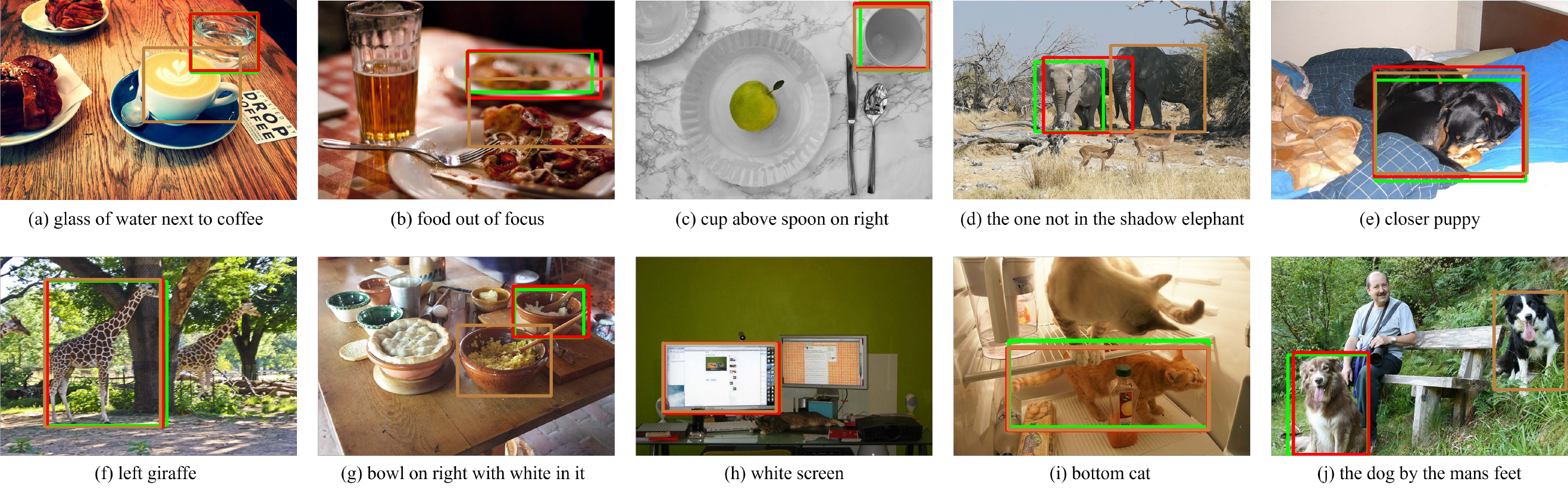}
    \caption{Qualitative results of our model compared with the state-of-the-art method VLTVG in the standard scene that train on RefCOCO training set and test on RefCOCO testB set. \color{Green}{Green: ground truth}; \color{Red}{Red: ours}; \color{Brown}{Brown: VLTVG}.}
    \label{fig:visualization_unc}
\end{figure*}

\begin{figure*}[t]
    \centering
    \includegraphics[width=1\linewidth]{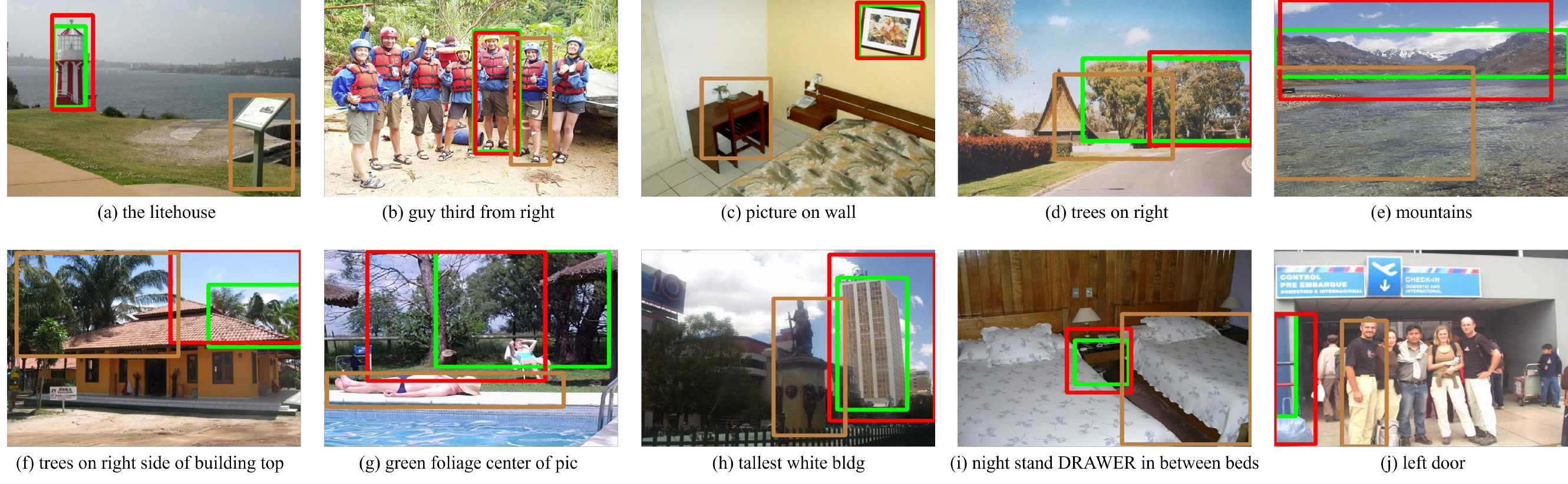}
    \caption{Qualitative results of our model compared with the state-of-the-art method VLTVG in the open-vocabulary scene that train on RefCOCO training set and test on ReferIt val set. \color{Green}{Green: ground truth}; \color{Red}{Red: ours}; \color{Brown}{Brown: VLTVG}.}
    \label{fig:visualization_open}
\end{figure*}

The qualitative results of the visual and language branches in the context disentangling module are presented in Fig.~\ref{fig:visualization_dis_score} and Fig.~\ref{fig:visualization_phrase_attn}. Fig.~\ref{fig:visualization_dis_score} shows the visualization of the discrimination coefficients, we can observe that all salient objects that appeared in the referring expressions are extracted with higher coefficients, while the context is suppressed by lower coefficients. Unlike visual disentangling, which enhances the salient objects, the phrase attention weights presented in Fig.~\ref{fig:visualization_phrase_attn} reveal that the context disentangling of the language branch focuses more attention on context information. For example, the queries "food out of focus" and "right dog" concentrates on "out of" and "right" which are the contextual relationships of the referents.

\subsection{Qualitative Comparisons}
Furthermore, in Fig.~\ref{fig:visualization_unc}, we show our qualitative comparison with the state-of-the-art method VLTVG. From Fig.~\ref{fig:visualization_unc}, we can observe that our model successfully grounds the object expressed in the given language expression even though the content is very difficult for machines to understand. For instance, in Fig.~\ref{fig:visualization_unc} (b), the context expression "out of focus" is a very challenging statement that requires the model to have a strong comprehension ability. Through context disentangling and prototype inheriting, our model accurately learns the essential features of the out-of-focus objects which may come from other training data, and then inherits them to distinguish the out-of-focus food from other foods in (b). The phenomenon can also be demonstrated in Fig.~\ref{fig:visualization_dis_score} and Fig.~\ref{fig:visualization_phrase_attn}. Additionally, it is worth noting that VLTVG fails to ground the referent in Fig.~\ref{fig:visualization_unc} (a), (b), (d), (g) and (j). Based on these failures of VLTVG, we hypothesize that it gives more attention to the larger object when there are multiple objects of the same category in the visual space. As a result, VLTVG fails to ground (a), (b), (d) and (g), but successfully locates the referent in (c), (e) (where there are no same-category objects), and (f), (h), (i) (where the referent is the larger object among same-category objects). 

The qualitative results of the open-vocabulary scene (trained on RefCOCO and tested on ReferIt val set) are presented in Fig.~\ref{fig:visualization_open}. Based on the results, it can be observed that our model benefits from the use of cluster-level prototypes. For instance, even when presented with incorrect or novel referent words, such as 'the litehouse' in Fig.~\ref{fig:visualization_open} (a) and 'night stand DRAWER' in (i), our model is still able to correctly identify and ground the referents. Other cases, such as those in (e), (g), and (h), demonstrate that TransCP correctly locates the positions of 'mountains', 'foliage', and 'bldg' (novel objects), whereas VLTVG draws incorrect bounding boxes around the water, person, and statue. Examples (b), (d), and (f) show that our model, with context disentangling, better distinguishes the referent from the context within the same categories. 

From the two qualitative comparisons, we can see that our model is more robust and can better handle both standard and open-vocabulary scenes for visual grounding than existing methods.

\subsection{Failure Cases Study}
\begin{figure*}[t]
    \centering
    \includegraphics[width=1\linewidth]{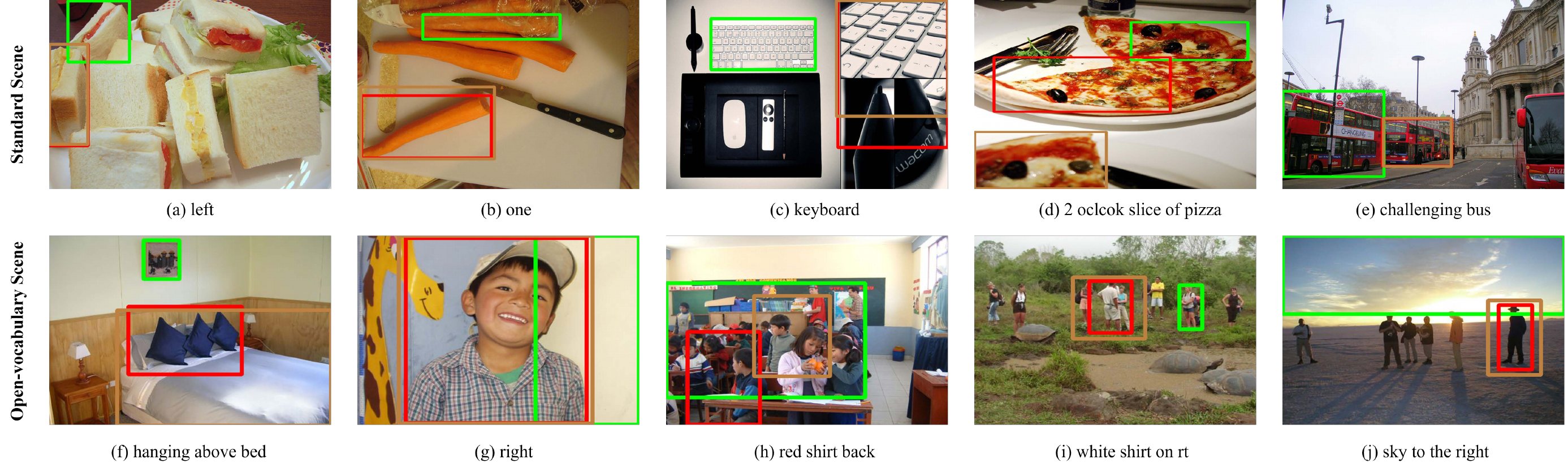}
    \caption{Some failure cases in the standard and open-vocabulary scenes. \color{Green}{Green: ground truth}; \color{Red}{Red: ours}; \color{Brown}{Brown: VLTVG}.}
    \label{fig:visualization_failure}
\end{figure*}
{To better analyze TransCP, some failure cases from the standard and the open-vocabulary scenes are presented in Fig.~\ref{fig:visualization_failure}.
From Fig.~\ref{fig:visualization_failure}, we can draw several observations.}
{First, when the referring expression lacks a subject or a clear referent, both models may fail because they both rely on enhancing salient objects, as shown in (a), (b), (f), and (g). 
Second, if the referent is highly ambiguous and the expression lacks discriminative information to distinguish the true referent, both TransCP and VLTVG fail to align with the ground truth. This is because visual grounding datasets typically provide only one region as the ground truth, while the given referring expression corresponds to multiple regions in the image, as seen with examples like "keyboard" in (c) or "red shirt" in (h).
Third, while TransCP learns prototypes on disentangled visual features to aid in open-vocabulary scenes (as seen in successful cases shown in Fig.~\ref{fig:visualization_open} (a), (h) and (i)), it may fail to transfer the rare (or misspelled) relations or attributes in the disentangled language features to the open-vocabulary scene (as seen with the "2 oclock" in (d) and "rt" in (i)).}
  
{There exist other failure cases. For instance, both models have limited OCR (Optical Character Recognition) capabilities, as shown in Fig.~\ref{fig:visualization_failure} (e). Additionally, if the semantics are uncommon in visual features of the training data, the prototype discovering module may not learn suitable prototypes to be inherited for open-vocabulary scenes (TransCP fails to transfer the broad prototype of "sky" from the training data in Fig.~\ref{fig:visualization_failure} (j)).}

\section{Conclusion and Limitations}

In this paper, we propose a novel framework, TransCP, based on Transformers with context disentangling and prototype inheriting. It can handle both standard and open-vocabulary scenes for robust visual grounding. Our context disentangling module enhances the discrimination between the referent and the context both in visual and linguistic features. The prototype bank discovers the prototypes during training and stores them for inheritance during inference to guide visual grounding, especially for the open-vocabulary scene. Moreover, the parameter-free Hadamard fusion strategy maintains the equal importance of visual and linguistic features, which benefits the robustness. Extensive experiments in both scenes exhibit the superiority of our method.

{The main limitation of our work is that our model's absolute performance in the open-vocabulary scene is still unsatisfactory for real-world applications. Additionally, it is challenging to handle extremely complex referring expressions for recent visual grounding methods. In future, we will integrate it with VLP models to improve the performance, while adapt it to incorporate reasoning methods to handle the more complex referring expressions.}
	
\section{Acknowledgments}
This work was supported by the National Key Research and Development Program of China (Grant 2022ZD0118802), the National Natural Science Foundation of China (Grant No. U20B2064, 62102181 and 62322211), and Youth Innovation Promotion Association of CAS (Grant 2020108).

\bibliographystyle{IEEEtran}
\bibliography{refer}

\begin{thebibliography}{10}
\providecommand{\url}[1]{#1}
\csname url@samestyle\endcsname
\providecommand{\newblock}{\relax}
\providecommand{\bibinfo}[2]{#2}
\providecommand{\BIBentrySTDinterwordspacing}{\spaceskip=0pt\relax}
\providecommand{\BIBentryALTinterwordstretchfactor}{4}
\providecommand{\BIBentryALTinterwordspacing}{\spaceskip=\fontdimen2\font plus
\BIBentryALTinterwordstretchfactor\fontdimen3\font minus \fontdimen4\font\relax}
\providecommand{\BIBforeignlanguage}[2]{{%
\expandafter\ifx\csname l@#1\endcsname\relax
\typeout{** WARNING: IEEEtran.bst: No hyphenation pattern has been}%
\typeout{** loaded for the language `#1'. Using the pattern for}%
\typeout{** the default language instead.}%
\else
\language=\csname l@#1\endcsname
\fi
#2}}
\providecommand{\BIBdecl}{\relax}
\BIBdecl

\bibitem{liu2019adaptive}
X.~Liu, L.~Li, S.~Wang, Z.-J. Zha, D.~Meng, and Q.~Huang, ``Adaptive reconstruction network for weakly supervised referring expression grounding,'' in \emph{Proceedings of IEEE/CVF International Conference on Computer Vision}, 2019, pp. 2611--2620.

\bibitem{yang2022improving}
L.~Yang, Y.~Xu, C.~Yuan, W.~Liu, B.~Li, and W.~Hu, ``Improving visual grounding with visual-linguistic verification and iterative reasoning,'' in \emph{Proceedings of IEEE/CVF Conference on Computer Vision and Pattern Recognition}, 2022, pp. 9499--9508.

\bibitem{chen2022understanding}
Y.-W. Chen, Y.-H. Tsai, and M.-H. Yang, ``Understanding synonymous referring expressions via contrastive features,'' \emph{International Journal of Computer Vision}, vol. 130, no.~10, pp. 2501--2516, 2022.

\bibitem{du2022visual}
Y.~Du, Z.~Fu, Q.~Liu, and Y.~Wang, ``Visual grounding with transformers,'' in \emph{Proceedings of IEEE International Conference on Multimedia and Expo}, 2022, pp. 1--6.

\bibitem{liu2022entity}
X.~Liu, L.~Li, S.~Wang, Z.-J. Zha, Z.~Li, Q.~Tian, and Q.~Huang, ``Entity-enhanced adaptive reconstruction network for weakly supervised referring expression grounding,'' \emph{IEEE Trans. Pattern Anal. Mach. Intell.}, vol.~45, no.~3, pp. 3003--3018, 2023.

\bibitem{HuangHGQZ19}
P.~Huang, J.~Huang, Y.~Guo, M.~Qiao, and Y.~Zhu, ``Multi-grained attention with object-level grounding for visual question answering,'' in \emph{Proceedings of AAAI Conference on Artificial Intelligence}, 2019, pp. 3595--3600.

\bibitem{LiTIP17}
Z.~Li and J.~Tang, ``Weakly supervised deep matrix factorization for social image understanding,'' \emph{IEEE Trans. Image Process.}, vol.~26, no.~1, pp. 276--288, 2017.

\bibitem{yu2020language}
D.~Rozenberszki, O.~Litany, and A.~Dai, ``Language-grounded indoor 3d semantic segmentation in the wild,'' in \emph{Proceedings of European Conference on Computer Vision}, vol. 13693, 2022, pp. 125--141.

\bibitem{CMN2017}
R.~Hu, M.~Rohrbach, J.~Andreas, T.~Darrell, and K.~Saenko, ``Modeling relationships in referential expressions with compositional modular networks,'' in \emph{Proceedings of IEEE/CVF Conference on Computer Vision and Pattern Recognition}, 2017, pp. 4418--4427.

\bibitem{VC2018}
H.~Zhang, Y.~Niu, and S.~Chang, ``Grounding referring expressions in images by variational context,'' in \emph{Proceedings of IEEE/CVF Conference on Computer Vision and Pattern Recognition}, 2018, pp. 4158--4166.

\bibitem{ParalAttn2018}
B.~Zhuang, Q.~Wu, C.~Shen, I.~D. Reid, and A.~van~den Hengel, ``Parallel attention: {A} unified framework for visual object discovery through dialogs and queries,'' in \emph{Proceedings of IEEE/CVF Conference on Computer Vision and Pattern Recognition}, 2018, pp. 4252--4261.

\bibitem{yu2018mattnet}
L.~Yu, Z.~Lin, X.~Shen, J.~Yang, X.~Lu, M.~Bansal, and T.~L. Berg, ``Mattnet: Modular attention network for referring expression comprehension,'' in \emph{Proceedings of IEEE/CVF International Conference on Computer Vision}, 2018, pp. 1307--1315.

\bibitem{similar_net2019}
L.~Wang, Y.~Li, J.~Huang, and S.~Lazebnik, ``Learning two-branch neural networks for image-text matching tasks,'' \emph{IEEE Trans. Pattern Anal. Mach. Intell.}, vol.~41, no.~2, pp. 394--407, 2019.

\bibitem{chen2018real}
X.~Chen, L.~Ma, J.~Chen, Z.~Jie, W.~Liu, and J.~Luo, ``Real-time referring expression comprehension by single-stage grounding network,'' \emph{arXiv preprint arXiv:1812.03426}, 2018.

\bibitem{sadhu2019zero}
A.~Sadhu, K.~Chen, and R.~Nevatia, ``Zero-shot grounding of objects from natural language queries,'' in \emph{Proceedings of IEEE/CVF International Conference on Computer Vision}, 2019, pp. 4694--4703.

\bibitem{yang2019fast}
Z.~Yang, B.~Gong, L.~Wang, W.~Huang, D.~Yu, and J.~Luo, ``A fast and accurate one-stage approach to visual grounding,'' in \emph{Proceedings of IEEE/CVF International Conference on Computer Vision}, 2019, pp. 4683--4693.

\bibitem{liao2020real}
Y.~Liao, S.~Liu, G.~Li, F.~Wang, Y.~Chen, C.~Qian, and B.~Li, ``A real-time cross-modality correlation filtering method for referring expression comprehension,'' in \emph{Proceedings of IEEE/CVF Conference on Computer Vision and Pattern Recognition}, 2020, pp. 10\,880--10\,889.

\bibitem{yang2020improving}
Z.~Yang, T.~Chen, L.~Wang, and J.~Luo, ``Improving one-stage visual grounding by recursive sub-query construction,'' in \emph{Proceedings of European Conference on Computer Vision}, 2020, pp. 387--404.

\bibitem{huang2021look}
B.~Huang, D.~Lian, W.~Luo, and S.~Gao, ``Look before you leap: Learning landmark features for one-stage visual grounding,'' in \emph{Proceedings of IEEE/CVF Conference on Computer Vision and Pattern Recognition}, 2021, pp. 16\,888--16\,897.

\bibitem{zhu2022seqtr}
C.~Zhu, Y.~Zhou, Y.~Shen, G.~Luo, X.~Pan, M.~Lin, C.~Chen, L.~Cao, X.~Sun, and R.~Ji, ``Seqtr: {A} simple yet universal network for visual grounding,'' in \emph{Proceedings of European Conference on Computer Vision}, vol. 13695, 2022, pp. 598--615.

\bibitem{deng2021transvg}
J.~Deng, Z.~Yang, T.~Chen, W.~Zhou, and H.~Li, ``Transvg: End-to-end visual grounding with transformers,'' in \emph{Proceedings of IEEE/CVF International Conference on Computer Vision}, 2021, pp. 1769--1779.

\bibitem{YangLY19}
S.~Yang, G.~Li, and Y.~Yu, ``Cross-modal relationship inference for grounding referring expressions,'' in \emph{Proceedings of IEEE/CVF Conference on Computer Vision and Pattern Recognition}, 2019, pp. 4145--4154.

\bibitem{liu2020transferrable}
X.~Liu, L.~Li, S.~Wang, Z.-J. Zha, D.~Meng, and Q.~Huang, ``Transferrable referring expression grounding with concept transfer and context inheritance,'' in \emph{Proceedings of ACM International Conference on Multimedia}, 2020, pp. 3938--3946.

\bibitem{ShiSJZ22}
Z.~Shi, Y.~Shen, H.~Jin, and X.~Zhu, ``Improving zero-shot phrase grounding via reasoning on external knowledge and spatial relations,'' in \emph{Proceedings of AAAI Conference on Artificial Intelligence}, 2022, pp. 2253--2261.

\bibitem{MDETR_KamathSLSMC21}
A.~Kamath, M.~Singh, Y.~LeCun, G.~Synnaeve, I.~Misra, and N.~Carion, ``{MDETR} - modulated detection for end-to-end multi-modal understanding,'' in \emph{Proceedings of IEEE/CVF Conference on Computer Vision and Pattern Recognition}, 2021, pp. 1760--1770.

\bibitem{Unitab_YangGW000LW22}
Z.~Yang, Z.~Gan, J.~Wang, X.~Hu, F.~Ahmed, Z.~Liu, Y.~Lu, and L.~Wang, ``Unitab: Unifying text and box outputs for grounded vision-language modeling,'' in \emph{Proceedings of European Conference on Computer Vision}, vol. 13696, 2022, pp. 521--539.

\bibitem{WangYMLBLMZZY22}
P.~Wang, A.~Yang, R.~Men, J.~Lin, S.~Bai, Z.~Li, J.~Ma, C.~Zhou, J.~Zhou, and H.~Yang, ``{OFA:} unifying architectures, tasks, and modalities through a simple sequence-to-sequence learning framework,'' in \emph{International Conference on Machine Learning}, vol. 162, 2022, pp. 23\,318--23\,340.

\bibitem{rosch1978principles}
E.~Rosch, ``Principles of categorization,'' in \emph{Cognition and categorization}, 1978, pp. 27--48.

\bibitem{murphy2002big}
G.~Murphy, \emph{The big book of concepts}.\hskip 1em plus 0.5em minus 0.4em\relax MIT press, 2004.

\bibitem{xu2022attribute}
W.~Xu, Y.~Xian, J.~Wang, B.~Schiele, and Z.~Akata, ``Attribute prototype network for any-shot learning,'' \emph{International Journal of Computer Vision}, vol. 130, no.~7, pp. 1735--1753, 2022.

\bibitem{li2020transferrable}
A.~Li, Z.~Lu, J.~Guan, T.~Xiang, L.~Wang, and J.-R. Wen, ``Transferrable feature and projection learning with class hierarchy for zero-shot learning,'' \emph{International Journal of Computer Vision}, vol. 128, pp. 2810--2827, 2020.

\bibitem{nagaraja2016modeling}
V.~K. Nagaraja, V.~I. Morariu, and L.~S. Davis, ``Modeling context between objects for referring expression understanding,'' in \emph{Proceedings of European Conference on Computer Vision}, 2016, pp. 792--807.

\bibitem{mao2016generation}
J.~Mao, J.~Huang, A.~Toshev, O.~Camburu, A.~L. Yuille, and K.~Murphy, ``Generation and comprehension of unambiguous object descriptions,'' in \emph{Proceedings of IEEE/CVF International Conference on Computer Vision}, 2016, pp. 11--20.

\bibitem{luo2017comprehension}
R.~Luo and G.~Shakhnarovich, ``Comprehension-guided referring expressions,'' in \emph{Proceedings of IEEE/CVF International Conference on Computer Vision}, 2017, pp. 7102--7111.

\bibitem{li2013partial}
L.~Li, S.~Jiang, Z.-J. Zha, Z.~Wu, and Q.~Huang, ``Partial-duplicate image retrieval via saliency-guided visual matching,'' \emph{IEEE MultiMedia}, vol.~20, no.~3, pp. 13--23, 2013.

\bibitem{hu2016natural}
R.~Hu, H.~Xu, M.~Rohrbach, J.~Feng, K.~Saenko, and T.~Darrell, ``Natural language object retrieval,'' in \emph{Proceedings of IEEE/CVF International Conference on Computer Vision}, 2016, pp. 4555--4564.

\bibitem{lu2019vilbert}
J.~Lu, D.~Batra, D.~Parikh, and S.~Lee, ``Vilbert: Pretraining task-agnostic visiolinguistic representations for vision-and-language tasks,'' in \emph{Proceedings of Advances in Neural Information Processing Systems}, vol.~32, 2019.

\bibitem{su2019vl}
W.~Su, X.~Zhu, Y.~Cao, B.~Li, L.~Lu, F.~Wei, and J.~Dai, ``{VL-BERT:} pre-training of generic visual-linguistic representations,'' in \emph{Proceedings of International Conference on Learning Representations}, 2020.

\bibitem{LiTM19}
Z.~Li, J.~Tang, and T.~Mei, ``Deep collaborative embedding for social image understanding,'' \emph{IEEE Trans. Pattern Anal. Mach. Intell.}, vol.~41, no.~9, pp. 2070--2083, 2019.

\bibitem{YangLY20ref_reasoning}
S.~Yang, G.~Li, and Y.~Yu, ``Graph-structured referring expression reasoning in the wild,'' in \emph{Proceedings of IEEE/CVF Conference on Computer Vision and Pattern Recognition}, 2020, pp. 9949--9958.

\bibitem{ZhengWTZCWW20}
Y.~Zheng, Z.~Wen, M.~Tan, R.~Zeng, Q.~Chen, Y.~Wang, and Q.~Wu, ``Modular graph attention network for complex visual relational reasoning,'' in \emph{Proceedings of Asian Conference on Computer Vision}, vol. 12627, 2020, pp. 137--153.

\bibitem{HuRDS19}
R.~Hu, A.~Rohrbach, T.~Darrell, and K.~Saenko, ``Language-conditioned graph networks for relational reasoning,'' in \emph{Proceedings of IEEE/CVF International Conference on Computer Vision}, 2019, pp. 10\,293--10\,302.

\bibitem{DengWWHLT22ATT_A}
C.~Deng, Q.~Wu, Q.~Wu, F.~Hu, F.~Lyu, and M.~Tan, ``Visual grounding via accumulated attention,'' \emph{IEEE Trans. Pattern Anal. Mach. Intell.}, vol.~44, no.~3, pp. 1670--1684, 2022.

\bibitem{redmon2017yolo9000}
J.~Redmon and A.~Farhad, ``Yolo9000: Better, faster, stronger,'' in \emph{Proceedings of IEEE/CVF International Conference on Computer Vision}, 2017, pp. 7263--7271.

\bibitem{luo2020multi}
G.~Luo, Y.~Zhou, X.~Sun, L.~Cao, C.~Wu, C.~Deng, and R.~Ji, ``Multi-task collaborative network for joint referring expression comprehension and segmentation,'' in \emph{Proceedings of IEEE/CVF Conference on computer vision and pattern recognition}, 2020, pp. 10\,034--10\,043.

\bibitem{sun2021iterative}
M.~Sun, J.~Xiao, and E.~G. Lim, ``Iterative shrinking for referring expression grounding using deep reinforcement learning,'' in \emph{Proceedings of IEEE/CVF Conference on Computer Vision and Pattern Recognition}, 2021, pp. 14\,060--14\,069.

\bibitem{LiS21reftr}
M.~Li and L.~Sigal, ``Referring transformer: {A} one-step approach to multi-task visual grounding,'' in \emph{Proceedings of Advances in Neural Information Processing Systems}, 2021, pp. 19\,652--19\,664.

\bibitem{JiangLHSH22Pseudo_Q}
H.~Jiang, Y.~Lin, D.~Han, S.~Song, and G.~Huang, ``Pseudo-q: Generating pseudo language queries for visual grounding,'' in \emph{Proceedings of IEEE/CVF Conference on Computer Vision and Pattern Recognition}, 2022, pp. 15\,492--15\,502.

\bibitem{0007CGPYC21}
Z.~Chen, J.~Chen, Y.~Geng, J.~Z. Pan, Z.~Yuan, and H.~Chen, ``Zero-shot visual question answering using knowledge graph,'' in \emph{International Semantic Web Conference}, vol. 12922, 2021, pp. 146--162.

\bibitem{hu2021vivo}
X.~Hu, X.~Yin, K.~Lin, L.~Zhang, J.~Gao, L.~Wang, and Z.~Liu, ``Vivo: Visual vocabulary pre-training for novel object captioning,'' in \emph{Proceedings of AAAI Conference on Artificial Intelligence}, vol.~35, no.~2, 2021, pp. 1575--1583.

\bibitem{li2020oscar}
X.~Li, X.~Yin, C.~Li, P.~Zhang, X.~Hu, L.~Zhang, L.~Wang, H.~Hu, L.~Dong, F.~Wei, Y.~Choi, and J.~Gao, ``Oscar: Object-semantics aligned pre-training for vision-language tasks,'' in \emph{Proceedings of European Conference on Computer Vision}, vol. 12375, 2020, pp. 121--137.

\bibitem{zareian2021open}
A.~Zareian, K.~D. Rosa, D.~H. Hu, and S.-F. Chang, ``Open-vocabulary object detection using captions,'' in \emph{Proceedings of IEEE/CVF Conference on Computer Vision and Pattern Recognition}, 2021, pp. 14\,393--14\,402.

\bibitem{TANG2022108792}
H.~Tang, C.~Yuan, Z.~Li, and J.~Tang, ``Learning attention-guided pyramidal features for few-shot fine-grained recognition,'' \emph{Pattern Recognition}, vol. 130, p. 108792, 2022.

\bibitem{qian2022multimodal}
R.~Qian, Y.~Li, Z.~Xu, M.-H. Yang, S.~Belongie, and Y.~Cui, ``Multimodal open-vocabulary video classification via pre-trained vision and language models,'' \emph{arXiv preprint arXiv:2207.07646}, 2022.

\bibitem{TangLPT20}
H.~Tang, Z.~Li, Z.~Peng, and J.~Tang, ``Blockmix: Meta regularization and self-calibrated inference for metric-based meta-learning,'' in \emph{Proceedings of ACM International Conference on Multimedia}, 2020, pp. 610--618.

\bibitem{huang2021soho}
Z.~Huang, Z.~Zeng, Y.~Huang, B.~Liu, D.~Fu, and J.~Fu, ``Seeing out of the box: End-to-end pre-training for vision-language representation learning,'' in \emph{Proceedings of IEEE/CVF Conference on Computer Vision and Pattern Recognition}, 2021, pp. 12\,976--12\,985.

\bibitem{van2017neural}
A.~V.~D. Oord, O.~Vinyals, and K.~Kavukcuoglu, ``Neural discrete representation learning,'' in \emph{Proceedings of Advances in Neural Information Processing Systems}, 2017, pp. 6306--6315.

\bibitem{A2020Effect}
A.~{\"{O}}zg{\"{u}}r and F.~Nar, ``Effect of dropout layer on classical regression problems,'' in \emph{Proceedings of Signal Processing and Communications Applications Conference}, 2020, pp. 1--4.

\bibitem{kazemzadeh2014referitgame}
S.~Kazemzadeh, V.~Ordonez, M.~Matten, and T.~Berg, ``Referitgame: Referring to objects in photographs of natural scenes,'' in \emph{Proceedings of Conference on Empirical Methods in Natural Language Processing}, 2014, pp. 787--798.

\bibitem{PlummerWCCHL17}
B.~A. Plummer, L.~Wang, C.~M. Cervantes, J.~C. Caicedo, J.~Hockenmaier, and S.~Lazebnik, ``Flickr30k entities: Collecting region-to-phrase correspondences for richer image-to-sentence models,'' \emph{International Journal of Computer Vision}, vol. 123, no.~1, pp. 74--93, 2017.

\bibitem{lin2014microsoft}
T.-Y. Lin, M.~Maire, S.~Belongie, J.~Hays, P.~Perona, D.~Ramanan, P.~Doll{\'a}r, and C.~L. Zitnick, ``Microsoft coco: Common objects in context,'' in \emph{Proceedings of European Conference on Computer Vision}, 2014, pp. 740--755.

\bibitem{feng2021encoder}
G.~Feng, Z.~Hu, L.~Zhang, and H.~Lu, ``Encoder fusion network with co-attention embedding for referring image segmentation,'' in \emph{Proceedings of IEEE/CVF Conference on Computer Vision and Pattern Recognition}, 2021, pp. 15\,506--15\,515.

\bibitem{HudsonM19gqa}
D.~A. Hudson and C.~D. Manning, ``{GQA:} {A} new dataset for real-world visual reasoning and compositional question answering,'' in \emph{Proceedings of IEEE/CVF Conference on Computer Vision and Pattern Recognition}, 2019, pp. 6700--6709.

\bibitem{KrishnaZGJHKCKL17visualgenome}
R.~Krishna, Y.~Zhu, O.~Groth, J.~Johnson, K.~Hata, J.~Kravitz, S.~Chen, Y.~Kalantidis, L.~Li, D.~A. Shamma, M.~S. Bernstein, and L.~Fei{-}Fei, ``Visual genome: Connecting language and vision using crowdsourced dense image annotations,'' \emph{International Journal of Computer Vision}, vol. 123, no.~1, pp. 32--73, 2017.

\bibitem{carion2020end}
N.~Carion, F.~Massa, G.~Synnaeve, N.~Usunier, A.~Kirillov, and S.~Zagoruyko, ``End-to-end object detection with transformers,'' in \emph{Proceedings of European conference on computer vision}, 2020, pp. 213--229.

\bibitem{CITE2018}
B.~A. Plummer, P.~Kordas, M.~H. Kiapour, S.~Zheng, R.~Piramuthu, and S.~Lazebnik, ``Conditional image-text embedding networks,'' in \emph{Proceedings of European Conference on Computer Vision}, vol. 11216, 2018, pp. 258--274.

\bibitem{DDPN2018}
Z.~Yu, J.~Yu, C.~Xiang, Z.~Zhao, Q.~Tian, and D.~Tao, ``Rethinking diversified and discriminative proposal generation for visual grounding,'' in \emph{Proceedings of International Joint Conference on Artificial Intelligence}, 2018, pp. 1114--1120.

\bibitem{LGRANs2019}
P.~Wang, Q.~Wu, J.~Cao, C.~Shen, L.~Gao, and A.~van~den Hengel, ``Neighbourhood watch: Referring expression comprehension via language-guided graph attention networks,'' in \emph{Proceedings of IEEE/CVF Conference on Computer Vision and Pattern Recognition}, 2019, pp. 1960--1968.

\bibitem{DGA2019}
S.~Yang, G.~Li, and Y.~Yu, ``Dynamic graph attention for referring expression comprehension,'' in \emph{Proceedings of IEEE/CVF Conference on Computer Vision and Pattern Recognition}, 2019, pp. 4643--4652.

\bibitem{RvG_Tree2019}
R.~Hong, D.~Liu, X.~Mo, X.~He, and H.~Zhang, ``Learning to compose and reason with language tree structures for visual grounding,'' \emph{IEEE Trans. Pattern Anal. Mach. Intell.}, vol.~44, no.~2, pp. 684--696, 2019.

\bibitem{NMTree2019}
D.~Liu, H.~Zhang, Z.~Zha, and F.~Wu, ``Learning to assemble neural module tree networks for visual grounding,'' in \emph{Proceedings of IEEE/CVF Conference on Computer Vision and Pattern Recognition}, 2019, pp. 4672--4681.

\bibitem{Ref-NMS2021}
L.~Chen, W.~Ma, J.~Xiao, H.~Zhang, and S.~Chang, ``Ref-nms: Breaking proposal bottlenecks in two-stage referring expression grounding,'' in \emph{Proceedings of AAAI Conference on Artificial Intelligence}, 2021, pp. 1036--1044.

\end{thebibliography}

 \begin{IEEEbiography}[{\includegraphics[width=1in,height=1.25in,clip,keepaspectratio]{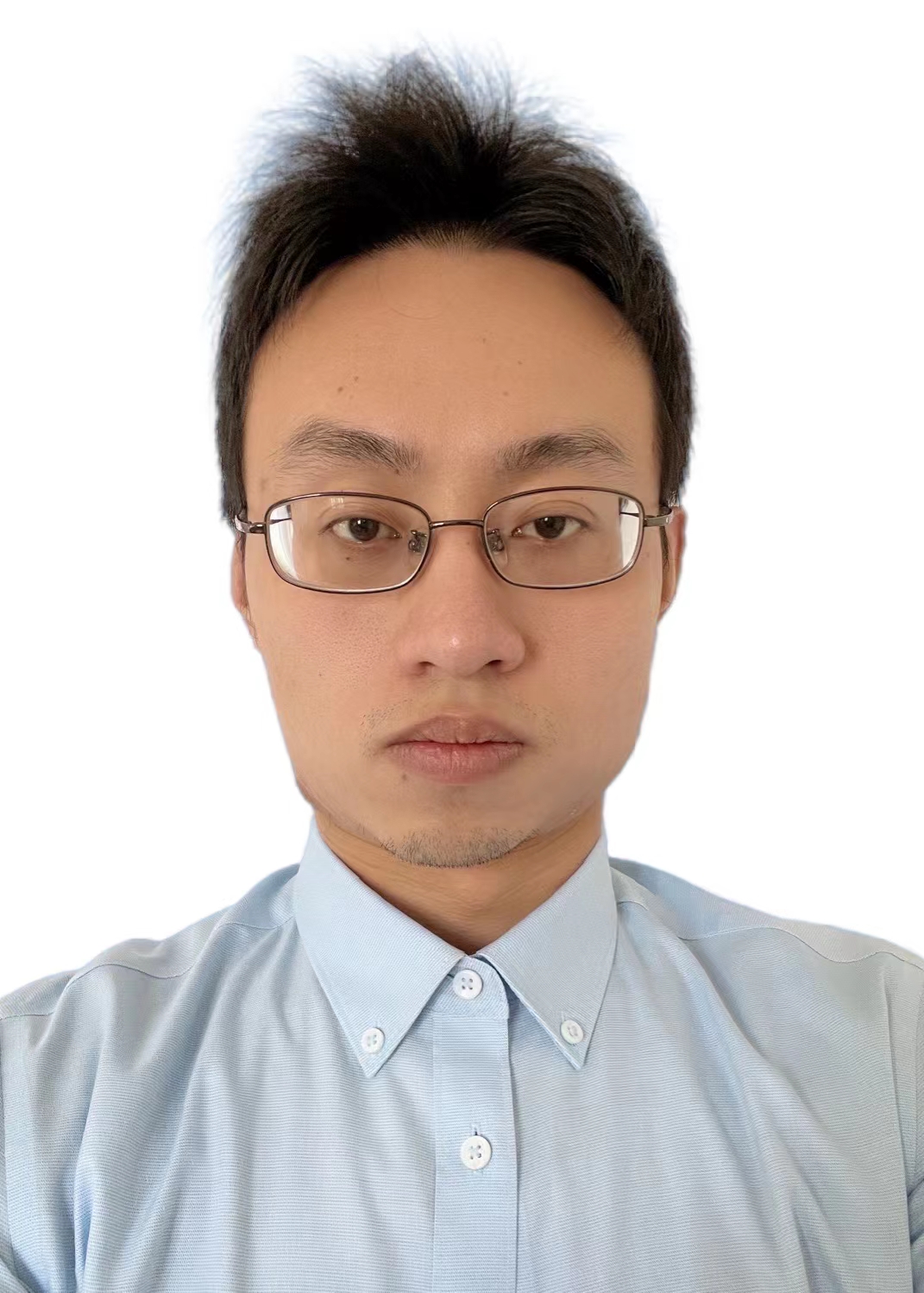}}] {Wei Tang}received the M.S. degree at Jiangsu University, China, in 2020. He is currently pursuing the Ph.D. degree with the School of Computer Science and Engineering, Nanjing University of Science and Technology, China. His research interests include deep learning, visual grounding and Vision-Language pre-training.
\end{IEEEbiography}

\begin{IEEEbiography}
[{\includegraphics[width=1in,height=1.25in,clip,keepaspectratio]{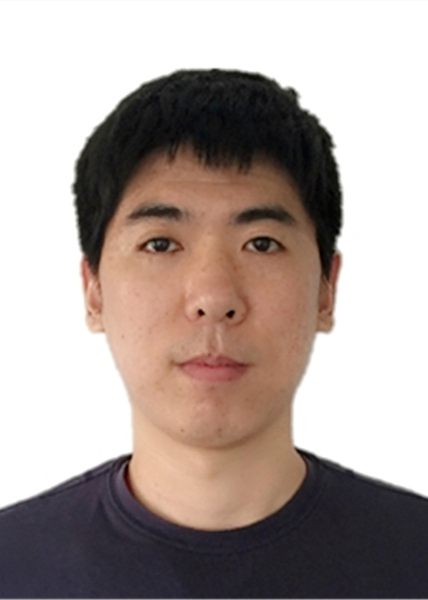}}] {Liang Li} received his B.S. degree from Xi’an Jiaotong Univerisity in 2008, and Ph.D. degree from Institute of Computing Technology, Chinese Academy of Sciences, Beijing, China in 2013. Currently he is serving as a associate professor at Institute of Computing Technology, Chinese Academy of Sciences. His research interests include image processing, large-scale image
retrieval, image semantic understanding, multimedia content analysis, computer vision, and pattern recognition.
\end{IEEEbiography}

\begin{IEEEbiography}
[{\includegraphics[width=1in,height=1.25in,clip,keepaspectratio]{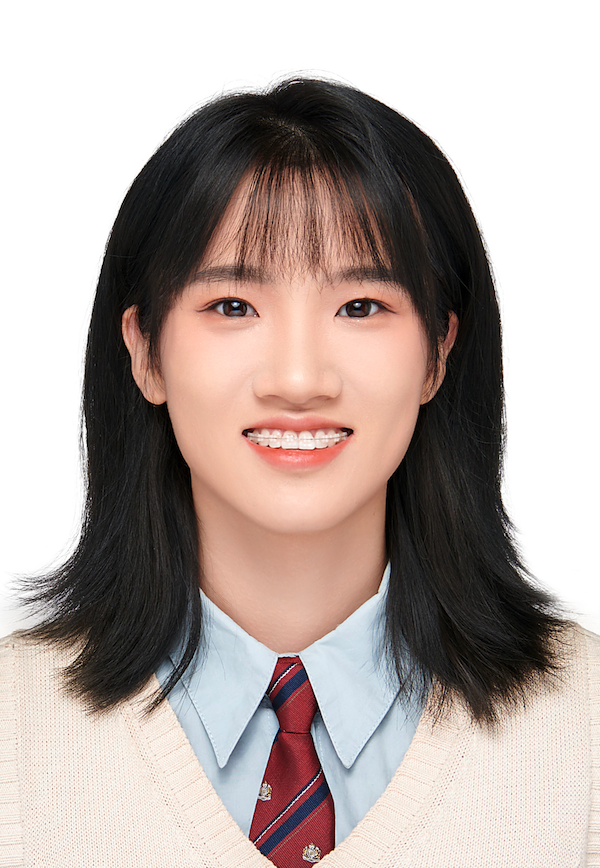}}] {Xuejing Liu} received her Ph.D. degree from the Institute of Computing Technology, Chinese Academy of Sciences. She holds a B.S. degree from Wuhan University. During her studies, she was affiliated with Key Laboratory of Intelligent Information Processing, Chinese Academy of Sciences. Her research interests include machine learning, deep learning, and computer vision. She is currently employed as an algorithm researcher at SenseTime Research.
\end{IEEEbiography}

\begin{IEEEbiography}
[{\includegraphics[width=1in,height=1.25in,clip,keepaspectratio]{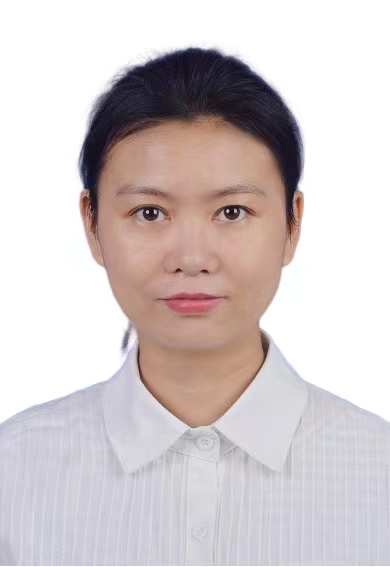  }}] {Lu Jin} is currently an Associate Professor at Nanjing University of Science and Technology, Nanjing, China. She received the Ph.D. degree from the Nanjing University of Science and Technology, Nanjing, China, in 2019. Her research interests include computer vision and multi-media retrieval.
\end{IEEEbiography}

\begin{IEEEbiography}[{\includegraphics[width=1in,height=1.25in,clip,keepaspectratio]{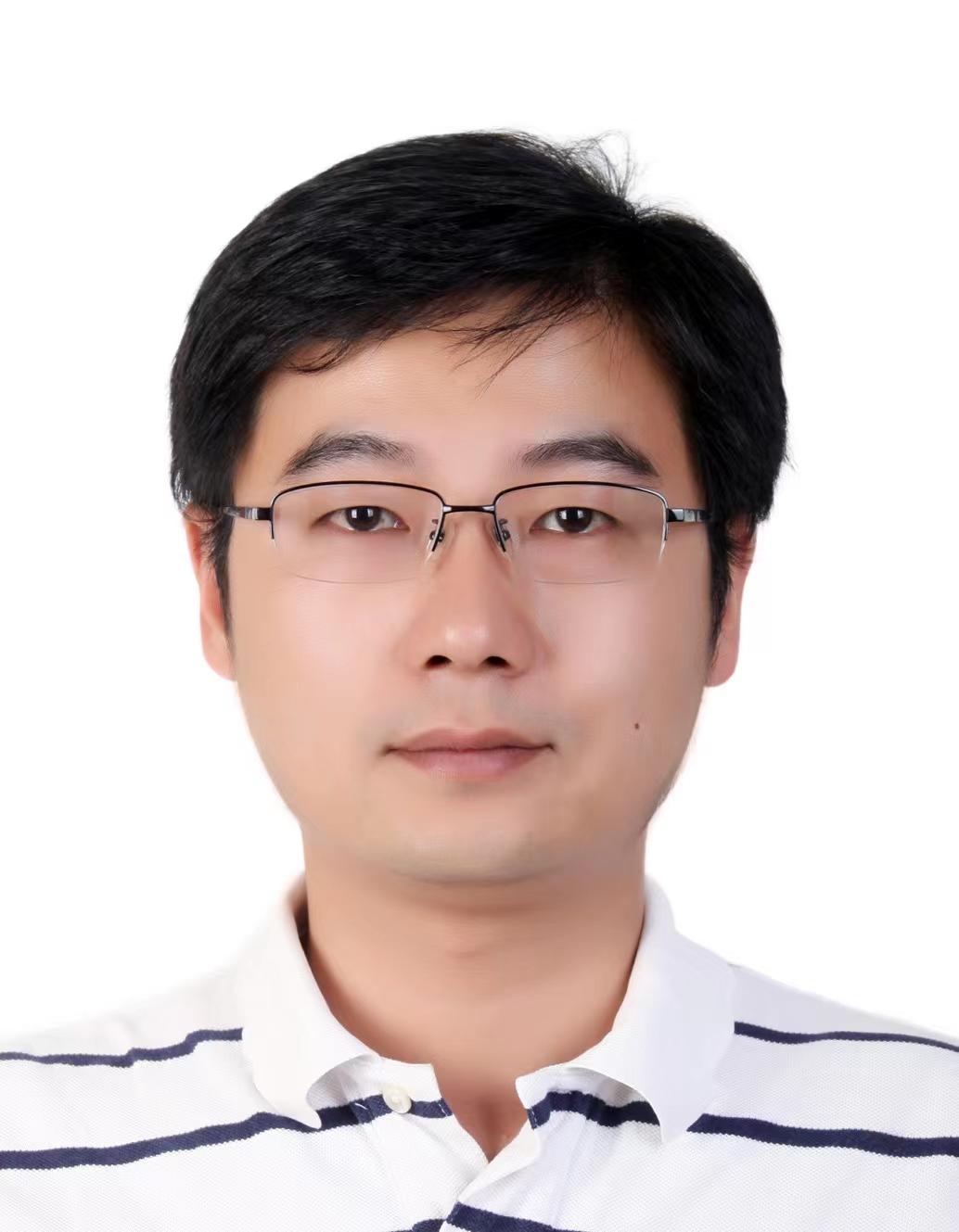}}] {Jinhui Tang} (Senior Member, IEEE) received the B.E. and Ph.D. degrees from the University of Science and Technology of China, Hefei, China, in 2003 and 2008, respectively. He is currently a Professor with the Nanjing University of Science and Technology, Nanjing, China. He has authored more than 200 articles in toptier journals and conferences. His research interests include multimedia analysis and computer vision. He was a recipient of the Best Paper Awards in ACM MM 2007 and ACM MM Asia 2020, the Best Paper Runner-Up in ACM MM 2015. He has served as an Associate Editor for IEEE TKDE, IEEE TMM, IEEE TNNLS and IEEE TCSVT. He is a Fellow of IAPR.
\end{IEEEbiography}

\begin{IEEEbiography}[{\includegraphics[width=1in,height=1.25in,clip,keepaspectratio]{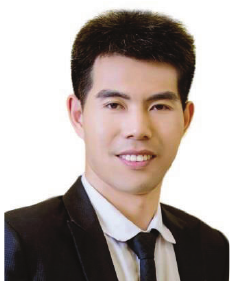}}] {Zechao Li} (Senior Member, IEEE) received the B.E. degree from the University of Science and Technology of China, Hefei, China, in 2008, and the Ph.D. degree from the National Laboratory of Pattern Recognition (PR), Institute of Automation, Chinese Academy of Sciences, Beijing, China, in 2013. He is currently a Professor with the Nanjing University of Science and Technology, Nanjing, China.
His research interests include big media analysis and computer vision. Dr. Li was a recipient of the Best Student Paper Award at ICIMCS 2018 and the Best Paper Award at the ACM Multimedia Asia 2020. He serves as an Associate Editor for the IEEE TNNLS and Information Sciences.
\end{IEEEbiography}

\end{document}